\documentclass[sigconf]{acmart}

\usepackage{graphicx}
\usepackage{amsmath}
\usepackage{amssymb}
\usepackage{booktabs}
\usepackage{multicol}
\usepackage{multirow}
\usepackage{amsthm}
\usepackage{mathrsfs}
\usepackage{diagbox}
\usepackage[ruled]{algorithm2e}
\usepackage{bm}
\usepackage{subfigure}
\usepackage[inline]{enumitem}

\newcommand{\system}{M2TR\xspace}

\newcommand*{\ie}{\emph{i.e.}\@\xspace}

\AtBeginDocument{%
  \providecommand\BibTeX{{%
    \normalfont B\kern-0.5em{\scshape i\kern-0.25em b}\kern-0.8em\TeX}}}

\copyrightyear{2022}
\acmYear{2022}
\setcopyright{acmcopyright}\acmConference[ICMR '22]{Proceedings of ACM
	International Conference on Multimedia Retrieval}{June 27--30, 2022}{Newark, NJ, USA}
\acmBooktitle{Proceedings of ACM International Conference on Multimedia Retrieval
	(ICMR '22), June 27--30, 2022, Newark, NJ, USA}

\begin{document}

\title{M2TR: Multi-modal Multi-scale Transformers for \\ Deepfake Detection}
   
\author{Junke Wang\textsuperscript{\rm 1,2}, Zuxuan Wu\textsuperscript{\rm 1,2}, Wenhao Ouyang\textsuperscript{\rm 1,2}, Xintong Han\textsuperscript{\rm 3} \\
Jingjing Chen\textsuperscript{\rm 1,2}, Ser-Nam Lim\textsuperscript{\rm 4}, Yu-Gang Jiang\textsuperscript{\rm 1,2  $\dagger$} \\
\textsuperscript{\rm 1}Shanghai Key Lab of Intelligent Information Processing, \\ School of Computer Science, Fudan University \\ 
\textsuperscript{\rm 2}Shanghai Collaborative Innovation Center on Intelligent Visual Computing,
 \textsuperscript{\rm 3}Huya Inc,  \textsuperscript{\rm 4}Meta AI \\
}
\thanks{$\dagger$ Corresponding author.}

\renewcommand{\shortauthors}{Wang, et al.}

\begin{abstract}
 The widespread dissemination of Deepfakes demands effective approaches that can detect perceptually convincing forged images. In this paper, we aim to capture the subtle manipulation artifacts at different scales using transformer models. In particular, we introduce a \textbf{M}ulti-modal \textbf{M}ulti-scale \textbf{TR}ansformer (\textbf{\system}), which operates on patches of different sizes to detect local inconsistencies in images at different spatial levels. \system further learns to detect forgery artifacts in the frequency domain to complement RGB information through a carefully designed cross modality fusion block. In addition, to stimulate Deepfake detection research, we introduce a high-quality Deepfake dataset, SR-DF, which consists of 4,000 DeepFake videos generated by state-of-the-art face swapping and facial reenactment methods. We conduct extensive experiments to verify the effectiveness of the proposed method, which outperforms state-of-the-art Deepfake detection methods by clear margins.
\end{abstract}


\maketitle

\section{Introduction}
\label{sec:intro}
Recent years have witnessed the rapid development of Deepfake techniques~\cite{kemelmacher2016transfiguring,koujan2020head2head,pumarola2018ganimation,suwajanakorn2017synthesizing}, which enable attackers to manipulate the facial regions of an image and generate a forged image. As the synthesized images are becoming more photo-realistic, it is extremely difficult to distinguish whether an image has been manipulated even for the human eyes. At the same time, these forged images might be distributed on the Internet for malicious purposes, which could bring societal implications. The above challenges have driven the development of Deepfake forensics using deep neural networks~\cite{zhou2017two, afchar2018mesonet, nguyen2019multi, li2019exposing, jeon2020fdftnet, li2020face, chen2021local}. Most existing approaches take as inputs a face region cropped out of an entire image and produce a binary real/fake prediction with deep CNN models ~\cite{simonyan2014very,he2016deep,chollet2017xception}. These methods capture artifacts from the face regions in a single scale with stacked convolutional operations~\cite{zhang2020resnest,tan2019efficientnet}. While decent detection results are achieved by stacked convolutions, they excel at modeling local information but fail to consider the relationships of pixels globally due to constrained receptive field.

\begin{figure}[t]
  \centering
  \includegraphics[width=\linewidth]{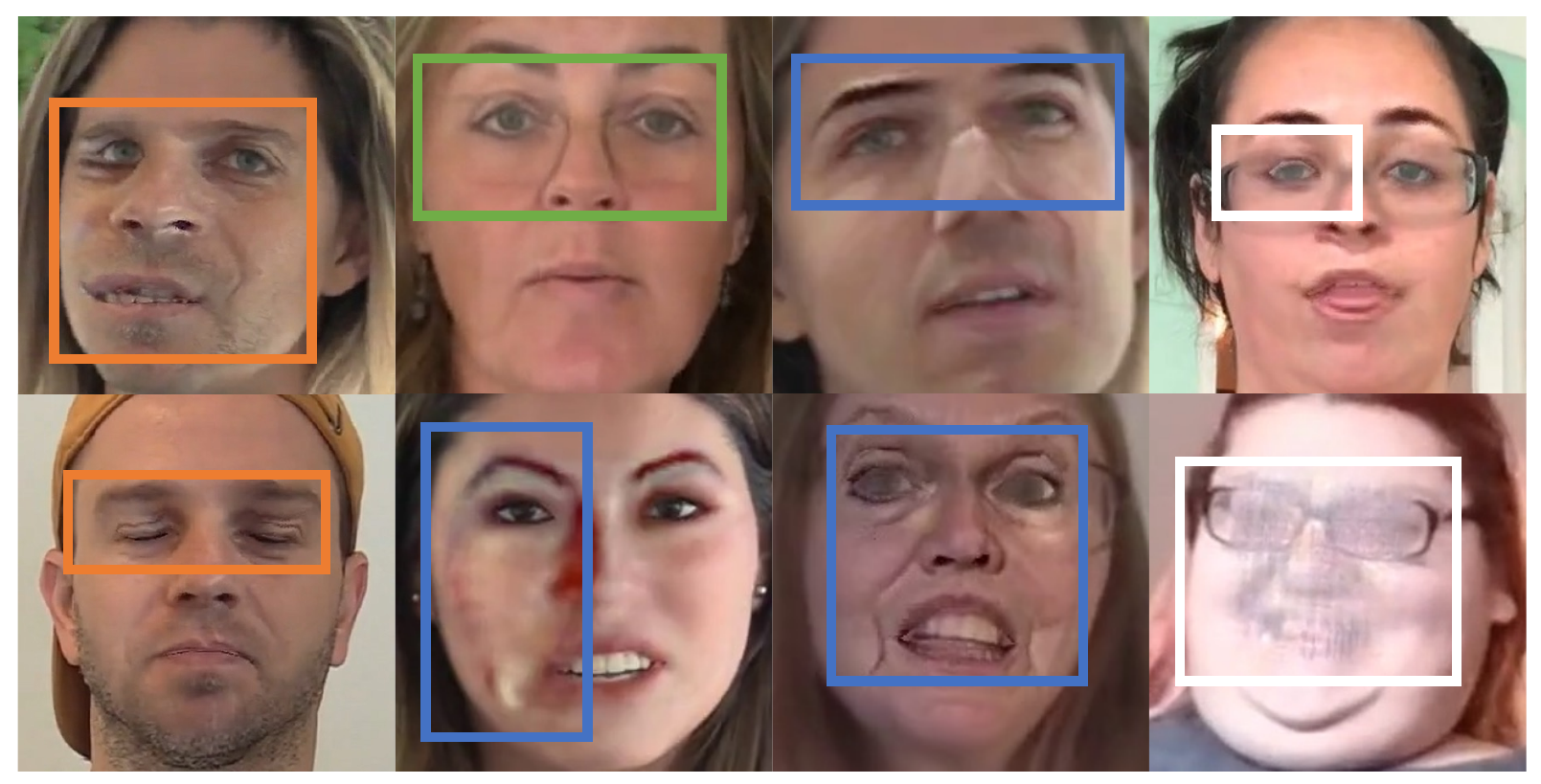}
  \vspace{-0.1in}
  \caption{Visual artifacts of images in the DFDC dataset~\cite{dolhansky2020deepfake}, including color mismatch (blue), shape distortion (orange), visible boundaries (green), and facial blurring (white).}
  \label{fig:bad}
\end{figure}

We posit that relationships among pixels are particularly useful for Deepfake detection, since pixels in certain artifacts are clearly different from the remaining pixels in the image. On the other hand, we observe that forgery patterns vary in size. For instance, Figure~\ref{fig:bad} gives examples from the DFDC dataset~\cite{dolhansky2020deepfake}. We can see that some forgery traces such as color mismatch occur in small regions (like the mouth corners), while 
 other forgery signals such as visible boundaries that almost span the entire image. Therefore, how to effectively explore regions of different scales in images is extremely critical for Deepfake detection. 
 
To address the above limitations, we explore transformers to model the relationships of pixels due to their strong capability of long-term dependency modeling for both natural language processing tasks~\cite{vaswani2017attention, devlin2018bert, raffel2019exploring} and computer vision tasks~\cite{dosovitskiy2020image, carion2020end, zhu2020deformable, wang2022objectformer, wang2021efficient}. Unlike traditional vision transformers that usually operate on a single-scale, we propose a multi-scale architecture to capture forged regions that potentially have different sizes. Furthermore, ~\cite{durall2019unmasking, qian2020thinking, huang2020deep, wang2020cnn} suggest that the artifacts of forged images will be destroyed by perturbations such as JPEG compression, making them imperceptible in the RGB domain but can still be detected in the frequency domain. This motivates us to use frequency information as a complementary modality in order to reveal artifacts that are no longer perceptible in the RGB domain.

\begin{figure*}[t]
  \centering
  \includegraphics[width=\linewidth]{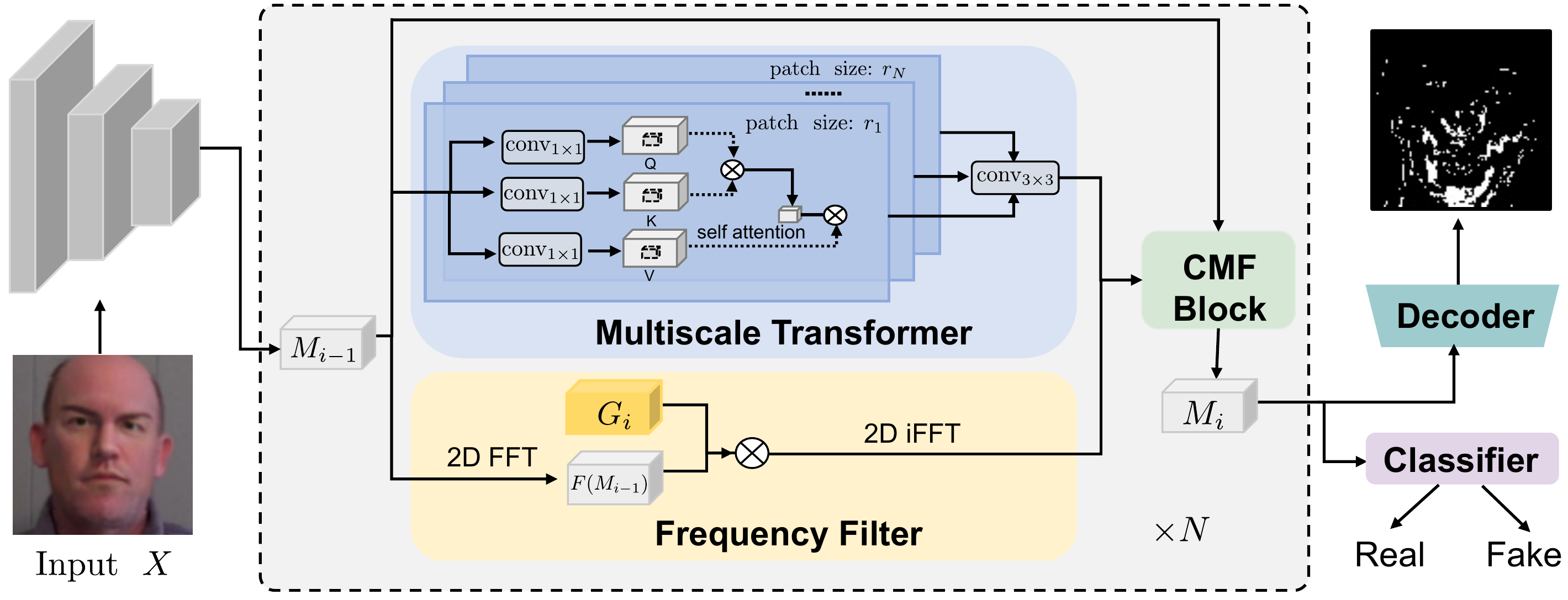}
  \vspace{-0.1in}
  \caption{Overview of the proposed \system. The input is a suspicious face image (H $\times$ W $\times$ C), and the output includes both a forgery detection result and a predicted mask (H $\times$ W $\times$ 1), which locates the forgery regions.}
  \label{fig:network}
\end{figure*}

To this end, we introduce~\system, a Multi-modal Multi-scale Transformer, for Deepfake detection. As illustrated in Figure~\ref{fig:network}, \system follows a two-stream architecture, where the RGB stream captures the inconsistency among different regions within an image at multiple scales in RGB domain, and the frequency stream adopts learnable frequency filters to filter out forgery features in frequency domain. We also design a cross modality fusion block to combine the information from both streams more effectively in an interactive fashion. Finally, the integrated features are input to fully connected layers to generate prediction results. In addition to binary classification, we also predict the manipulated regions of the face image in a multi-task manner. The rationale behind is that binary classification tends to result in easily overfitted models. Therefore, we use face masks as additional supervisory signals to mitigate overfitting. 

The availability of large-scale training data is an essential factor in the development of Deepfake detection methods. However, the quality of visual samples in current Deepfake datasets~\cite{yang2019exposing,korshunov2018deepfakes,rossler2019faceforensics++,dolhansky2020deepfake,li2020celeb} is limited, containing clear artifacts (see Figure~\ref{fig:bad}) like color mismatch, shape distortion, visible boundaries, and facial blurring. Therefore, there is still a huge gap between the images in existing datasets and forged images in the wild which are circulated on the Internet. Although the visual quality of Celeb-DF~\cite{li2020celeb} is relatively high compared to others, they use only one face swapping method to generate forged images, lacking sample diversity. In addition, there are no unbiased and comprehensive evaluation metrics to measure the quality of Deepfake datasets, which is not conducive to the development of subsequent Deepfake research.

In this paper, we present a large-scale and high-quality Deepfake dataset, \textbf{S}wapping and \textbf{R}eenactment \textbf{D}eep\textbf{F}ake (\textbf{SR-DF}) dataset, which is generated using the state-of-the-art face swapping and facial reenactment methods ~\cite{nirkin2019fsgan,faceshifter,siarohin2020first,tripathy2020icface} for the development and evaluation of Deepfake detection methods. Besides, we propose a set of evaluation criteria to measure the quality of Deepfake datasets from different perspectives.  We hope the release of SR-DF dataset and the evaluation systems will benefit the future research of Deepfake detection. Our work makes the following key contributions:

\begin{itemize}[leftmargin=*]
\item We propose a Multi-modal Multi-scale Transformer (\system) for Deepfake forensics, which uses a multi-scale transformer to detect local inconsistencies at different scales and leverages frequency features to improve the robustness. Extensive experiments demonstrate that our method achieves state-of-the-art detection performance on different datasets.
\item We introduce a large-scale and challenging Deepfake dataset SR-DF, which is generated with state-of-the-art face swapping and facial reenactment methods 
\item We construct the most comprehensive evaluation system and demonstrate that SR-DF dataset is well-suited for training Deepfake detection methods due to its quality and diversity.
\end{itemize}

\section{Related Work}
\noindent \textbf{Deepfake Detection} To mitigate the security threat brought by Deepfakes, a variety of methods have been proposed for Deepfake detection.~\cite{zhou2017two} uses a two-stream architecture to capture facial manipulation clues and patch inconsistency separately, while~\cite{nguyen2019multi} simultaneously identifies forged faces and locates the manipulated regions with multi-task learning. 

Recently, Face X-ray~\cite{li2020face} proposes to detect the blending boundaries based on an observation that the step of blending a forged face into the background is commonly used by most existing face manipulation methods. DCViT~\cite{wodajo2021deepfake} extracts features from the face image using a CNN model, which are then fed to a traditional single-scale transformer for forgery detection. MaDD~\cite{zhao2021multi} proposes a multi-attentional Deepfake detection framework to capture artifacts with multiple attention maps. However, most of them only focus on the features in the RGB domain, thus failing to detect forged images which are manipulated subtly in the color-space. Instead, F$^{3}$-Net~\cite{qian2020thinking} adopts a two-branch architecture where one makes use of frequency clues to recognize forgery patterns and the other extracts the discrepancy of frequency statistics between real and fake images. In this paper, we use a multi-scale transformer to capture local inconsistencies at different scales for forgery detection, and additionally introduce frequency modality to improve the robustness of our method to various image compression algorithms.

\vspace{0.1in}
\noindent \textbf{Visual Transformers} 
Transformers~\cite{vaswani2017attention} have demonstrated impressive performance for natural language processing tasks due to strong abilities in modeling long-range context information. Recently, researchers have demonstrated remarkable interests in using the transformer for a variety of computer vision tasks. Typically, visual transformers~\cite{dosovitskiy2020image, touvron2020training, carion2020end} model the interactions between tokens of the same scale with the self-attention mechanisms. ViT reshapes an image into a sequence of flattened patches and inputs them to the transformer encoder for image classification~\cite{dosovitskiy2020image}. DETR uses a common CNN to extract semantic features from the input image, which are then input to a transformer-based encoder-decoder architecture for object detection~\cite{carion2020end}.  Unlike these approaches, we propose to split the inputs into patches of different sizes, and integrate multi-scale information for better visual representation with vision transformers.

\section{Approach}
We aim to detect the subtle forgery artifacts that are hidden in the forged images and improve the robustness to image compression with frequency features. In this section, we introduce the Multi-modal Multi-scale Transformer (\system) for Deepfake detection, which consists of stacked multi-scale transformers (Sec~\ref{sec:mt}), frequency filters (Sec~\ref{sec:ff}), and cross modality fusion blocks (Sec~\ref{sec:cmf}). Figure~\ref{fig:network} gives an overview of the framework.

More formally, we denote an input image as \textit{X} $\in$ $\mathbb{R}^{H \times W \times 3}$, where \textit{H} and \textit{W} are the height and width of the image, respectively. We first use several convolutional layers to extract features $F$ $\in \mathbb{R}^{(H/4) \times (W/4) \times C}$ of \textit{X}, which are then input to successive multi-scale transformer and frequency filters for forgery clues detection. The intuition of using convolutions here is to ensure faster convergence and more stable training~\cite{xiao2021early}.

\subsection{Multi-scale Transformer}
\label{sec:mt}

To capture forgery patterns at multiple scales, we introduce a multi-scale transformer which operates on patches of different sizes. Taking the output of the previous cross modality fusion block $M_{i-1}$ (which is initialized as \textit{F}) as input, we split it into spatial patches of different sizes and calculate patch-wise self-attention in different heads. Specifically, we first extract patches of shape $r_{h} \times r_{h} \times C$ from $M_{i-1}$, and reshape them into 1-dimension vectors for the $h$-th head. After that, we use fully-connected layers to embed the flattened vectors into query embeddings $Q_{i}^{h}$ $\in$ $\mathbb{R}^{N \times C_{h}}$, where $N = (H/4r_{h}) \times (W/4r_{h})$, and $C_{h}=r_{h} \times r_{h} \times C$.  Similar operations are implemented to obtain key embeddings $K_{i}^{h}$ and value embeddings $V_{i}^{h}$, respectively. Then we calculate the attention matrix through the following process:
\begin{equation}
    A_{i}^{h} = softmax \left(\frac{Q_{i}^{h} (K_{i}^{h})^{T}}{C_{h}} \right)V_{i}^{h},
\end{equation}
$A_{i}^{h}$ is then reshaped to the original spatial resolution. Finally, the features from different heads are concatenated and further passed through a 2D residual block to obtain the output $T_{i} \in \mathbb{R}^{(H/4) \times (W/4) \times C}$. 

\subsection{Frequency Filter}
\label{sec:ff}
It has been shown that artifacts in manipulated images and videos are no longer perceptible with compression approaches like JPEG compression~\cite{durall2019unmasking, huang2020deep, wang2020cnn}. Therefore, we extract the forgery features in the frequency domain to complement RGB features.

Specifically, we first apply 2D FFT along the spatial dimensions to transform $M_{i-1}$ to the frequency domain, and obtain the spectrum representation $\mathcal{F}(M_{i-1}) \in \mathbb{R}^{H/4 \times W/4 \times C}$. We then multiply $\mathcal{F}(M_{i-1})$ with a learnable filter $G_{i} \in \mathbb{R}^{H/4 \times W/4 \times C}$ to model the dependencies of different frequency band components:
\begin{equation}
\hat{G}_{i} = G_{i} \odot \mathcal{F}(M_{i-1}),
\end{equation}
where $\odot$ denotes the Hadamard product. Finally, we perform the inverse FFT to covert $\hat{G}_{i}$ back to the spatial domain and obtain frequency-aware features $W_{i}$.

\subsection{Cross Modality Fusion}
\label{sec:cmf}
Given RGB features $T_{i}$ and frequency features $W_{i}$, we use a Cross Modality Fusion (CMF) block to fuse them into a unified representation. Inspired by the architecture of self-attention in transformers, we design a fusion block using the query-key-value mechanism. 

Specifically, we first embed $T_{i}$ and $W_{i}$ into $Q$, $K$, and $V$ using $1 \times$ 1 convolutions $conv_q$, $conv_k$, and $conv_v$, respectively. Then we flatten them along the spatial dimension to obtain the 2D embeddings $\widetilde{Q}$, $\widetilde{K}$, and $\widetilde{V}$ $\in \mathbb{R}^{(HW/16) \times C}$, and calculate the fused features as:
\begin{equation}
    \widetilde{M}_{i} = softmax \left( \frac{\widetilde{Q} \widetilde{K}^{T}} {\sqrt{H/4 \times W/4 \times C}} \right)\widetilde{V}.
\end{equation}

Finally, we employ a residual connection by adding $\widetilde{M}_{i}$ and $T_{i}$, and use a $3 \times 3$ convolution to obtain the output $M_{i}$:
\begin{equation}
    M_{i} = conv_{3 \times 3} (\widetilde{M}_{i} + T_{i}),
\end{equation}
where $M_{i} \in \mathbb{R}^{H/4 \times W/4 \times C}$ combines the forgery features in both RGB domain and frequency domain.

We stack the multi-scale transformer, frequency filter and CMF block for $N$ times ($N = 4$ in this paper). The integrated features $M_{out}$ are calculated by iterative update. Finally, we pass $M_{out}$ through several convolutional layers to obtain global semantic features $f$. 

\subsection{Loss functions}
\noindent \textbf{Cross-entropy loss}. We input $f$ to several fully-connected layers to predict whether the input image is real or fake using a cross-entropy loss $\mathcal{L}_{cls}$:
\begin{align}
    \mathcal{L}_{cls} = ylog\hat{y} + (1-y)log(1-\hat{y}),
    \label{eqn:cls}
\end{align}
where $y$ is set to 1 if the face image has been manipulated, otherwise it is set to 0; $\hat{y}$ denotes the predicted label by our network.

\vspace{0.1in}
\noindent \textbf{Segmentation loss} It is worth noting using a binary classifier tends to result in overfitted models. We additionally predict the face region as an auxiliary task to enrich the supervision for training the networks. Specifically, we input the feature map $M_{out}$ to a decoder (stacked convolutional layers and interpolation upsampling layers) to produce a binary mask $\bm{\hat{M}}$ $in$ $\mathbb{R}^{H \times W}$:
\begin{align}
    \mathcal{L}_{seg} = \sum_{i,j}M_{i,j}log\hat{M}_{i,j} + (1-M_{i,j})log(1-\hat{M}_{i,j}),
     \label{eqn:seg}
\end{align}
where $M_{i,j}$ is the ground-truth mask, with 1 indicating the manipulated pixels and 0 otherwise. 

\vspace{0.1in}
\noindent \textbf{Contrastive loss} Deepfake images generated by different facial manipulation methods differ in forgery patterns, while the distribution of real images is relatively stable. To improve the generalization ability of our detection model, we first calculate the feature centers of $N_p$ real samples $\bm{C}_{pos} = \frac{1}{N_p} \sum_{i=1}^{N_p} f_i^{pos}$ and additionally use a contrastive loss to make features from pristine samples to be closer towards the feature center than manipulated samples. Formally, the contrastive loss is defined as:
\begin{align}
    L_{con} = \frac{1}{N_p} \sum_{i=1}^{N_p} d(\bm{f}_{i}^{pos}, \bm{C}_{pos}) - \frac{1}{N_n} \sum_{i=1}^{N_n} d(\bm{f}_{i}^{neg}, \bm{C}_{pos}),
    \label{eqn:con}
\end{align}
where $N_n$ denotes the number of negative samples, and $d$ computes distance with cosine similarity. Finally, combining Eqn.~\ref{eqn:cls}, Eqn.~\ref{eqn:seg} and Eqn.~\ref{eqn:con}, the training objective can be written as:
\begin{equation}
    \mathcal{L} = \mathcal{L}_{cls} + \lambda_{1}\mathcal{L}_{seg} + \lambda_{2}\mathcal{L}_{con},
\end{equation}
where $\lambda_{1}$ and $\lambda_{2}$ are the balancing hyperparameters. By default, we set $\lambda_{1} = 1$ and $\lambda_{2} = 0.001$.

\section{SR-DF DATASET}
To stimulate research for Deepfake forensics, we introduce a large-scale and challenging dataset, SR-DF. SR-DF is built upon the pristine videos in the FF++ dataset, which contains a diverse set of samples in different genders, ages, and ethnic groups. We first crop face regions in each video frame using~\cite{king2009dlib}, and then generate forged videos using state-of-the-art Deepfake generation techniques. Finally, we use the image harmonization method in~\cite{cong2020dovenet} for post-processing. Below, we introduce these steps in detail.

\subsection{Dataset Construction}
\noindent \textbf{Synthesis Approaches} To guarantee the diversity of synthesized images, we use four facial manipulation methods, including two face swapping methods: \textbf{FSGAN}~\cite{nirkin2019fsgan} and \textbf{FaceShifter}~\cite{faceshifter}, and two facial reenactment methods: \textbf{First-order-motion}~\cite{siarohin2020first} and \textbf{IcFace}~\cite{tripathy2020icface}.  Note that the manipulation methods we leverage are all identity-agnostic---they can be applied to arbitrary face images without training in pairs, which is different from the FF++~\cite{rossler2019faceforensics++} dataset. The detailed forgery images generation process is described below.

\vspace{0.1in}
\noindent \textbf{\textit{FSGAN}}~\cite{nirkin2019fsgan} follows the following pipeline to swap the faces of the source image $I_{s}$ to that of the target image $I_{t}$. First, the swap generator $G_{r}$ estimates the swapped face $I_{r}$ and its segmentation mask $S_{r}$ based on $I_{t}$ and a heatmap encoding the facial landmark of $I_{s}$, while $G_{s}$ estimates the segmentation mask $S_{s}$ of the source image $I_{s}$. Then the inpainting generator $G_{c}$ inpaints the missing parts of $I_{r}$ based on $S_{s}$ to estimate the complete swapped face $I{c}$. Finally, using the segmentation mask $S_{s}$ , the blending generator $G_{b}$ blends $I_{c}$ and $I_{s}$ to generate the final output $I_{b}$ which preserves the  posture of $I_{s}$ but owns the identity of $I_{t}$. For our dataset, we directly use the pretrained model provided by \cite{nirkin2019fsgan} and inference on our pristine videos.

\vspace{0.1in}
\noindent \textbf{\textit{FaceShifter}}~\cite{li2019faceshifter} consists of two networks for full pipeline: AEI-Net for face swapping, and HEAR-Net for occlusion handling. As the author of \cite{li2019faceshifter} have not public their code, we use the code from \cite{faceshifter} who only implements AEI-Net, and we train the model on our data. Specifically, AEI-Net is composed of three components: 1) an Identity Encoder which adopts a pretrained state-of-the-art face recognition model to provide representative identity embeddings. 2) a Multi-level Attributes Encoder which encodes the features of facial attributes. 3) an AAD-Generator which integrates the information of identity and attributes in multiple feature levels and generates the swapped faces. We use the parameters declared in \cite{li2019faceshifter} to train the model. 
\vspace{0.1in}
\noindent \textbf{\textit{First-order-motion}}~\cite{siarohin2020first} decouples appearance and motion information for subject-agnostic facial reenactment. Their framework consists of two main modules: the motion estimation module which uses a set of learned keypoints along with their local affine transformations to predict a dense motion field, and an image generation module which combines the appearance extracted from the source image and the motion derived from the driving video to model the occlusions arising during target motions. To process our dataset, we use the pretrained model on VoxCeleb dataset \cite{nagrani2017voxceleb}, which contains speech videos from speakers spanning a wide range of different ethnic groups, accents, professions and ages, and reenact the faces in our real videos.

\vspace{0.1in}
\noindent \textbf{\textit{IcFace}}~\cite{tripathy2020icface} is a generic face animator that is able to transfer the expressions from a driving image to a source image. Specifically, the generator $G_{N}$ takes the source image and neutral facial attributes as input and produces the source identity with central pose and neutral expression. Then the generator $G_{A}$ takes the neutral image and attributes extracted from the driving image as an input and produces an image with the source identity and driving image’s attributes. We train the complete model on our real videos in a self-supervised manner, using the parameters that they use to train on VoxCeleb dataset \cite{nagrani2017voxceleb}. 

\vspace{0.1in}
\noindent \textbf{Post-processing} In order to resolve the color mismatch between the face regions and the background and to eliminate the stitched boundaries, we use DoveNet~\cite{cong2020dovenet} for post-processing, which is a state-of-the-art image harmonization method to make the foreground compatible with the background. Note that the masks that we use to distinguish foreground and background are generated using a face parsing model~\cite{face-parsing}. 

\subsection{Comparisons to current Deepfake Datasets}
We summarize the basic information of these existing Deefake datasets and our SR-DF dataset in Table~\ref{table1}. In addition, as mentioned above, how to measure the quality of forged images in these datasets remains under-explored. Therefore, we introduce a variety of quantitative metrics to benchmark the quality of current datasets from four perspectives: identity retention, authenticity, temporal smoothness, and diversity. To the best of our knowledge, this is the most comprehensive evaluation system to measure the quality of Deepfake datasets.

\begin{table}[!ht]
\centering
  \caption{A comparison of SR-DF dataset with existing datasets for Deepfake detection. LQ: low-quality, HQ: high-quality.}
  \label{table1}
  \vspace{-0.1in}
  \setlength{\tabcolsep}{0pt} 
    \begin{tabular*}{\linewidth}{@{\extracolsep{\fill}}*{7}l@{}}
    \toprule
    \multirow{2}{*}{\textbf{Dataset}} && \multicolumn{2}{c}{\textbf{Real}} && \multicolumn{2}{c}{\textbf{Forged}} \\
    ~ && Video & Frame && Video & Frame \\
    \cmidrule{1-1} \cmidrule{3-4} \cmidrule{6-7}
    UADFV~\cite{yang2019exposing} && 49 & 17.3k && 49 & 17.3k \\
    DF-TIMIT-LQ~\cite{korshunov2018deepfakes} && 320 & 34.0k && 320 & 34.0k \\
    DF-TIMIT-HQ~\cite{korshunov2018deepfakes} && 320 & 34.0k && 320 & 34.0k \\
    FF++~\cite{rossler2019faceforensics++} && 1,000 & 509.9k && 4000 & 1,830.1k\\
    DFD~\cite{dfd} && 363 & 315.4k && 3,068 & 2,242.7k\\
    DFDC~\cite{dolhansky2020deepfake} && 1,131 & 488.4k && 4,113 & 1,783.3k\\
    WildDeepfake~\cite{zi2020wilddeepfake} && 3,805 & 440.5k && 3,509 & 739.6k \\
    Celeb-DF~\cite{li2020celeb}  && 590 & 225.4k && 5,639 & 2,116.8k\\
    ForgeryNet~\cite{he2021forgerynet} && 99.6k & 1,438.2k && 121.6k & 1457.9k \\
    \textbf{SR-DF (ours)} && 1,000 & 509.9k && 4,000 & 2,078.4k\\
  \bottomrule
\end{tabular*}
\end{table}

\vspace{0.1in}
\noindent \textbf{Mask-SSIM} First, we follow \cite{li2020celeb} to adopt the Mask-SSIM score as a measurement of synthesized Deepfake images. Mask-SSIM refers to the SSIM score between the face regions of the forged image and the corresponding original image. We use the FaceParsing\cite{face-parsing} to generate facial masks and compute the Mask-SSIM on our face swapping subsets. Table~\ref{table2} demonstrates the average Mask-SSIM scores of all compared datasets, and SR-DF dataset achieves the highest scores.

\begin{table}[!ht]
\centering
  \caption{Average Mask-SSIM scores, perceptual loss, and $E_{warp}$ values of different Deepfake datasets, with the higher value corresponding to better image quality. For perceptual loss, lower value indicates the better image quality, and for $E_{warp}$, lower value corresponding to smoother temporal results.}
  \label{table2}
  \vspace{-0.1in}
  \setlength{\tabcolsep}{0pt} 
    \begin{tabular*}{\linewidth}{@{\extracolsep{\fill}}lcccccc@{}}
    \toprule
    \textbf{Dataset} && FF++ & DFD & DFDC & Celeb-DF & Ours \\
    \cmidrule{1-1} \cmidrule{3-7}
    \textbf{Mask-SSIM} $\uparrow$ && 0.82 & 0.86 & 0.85 & 0.91 &\textbf{0.92} \\
    \textbf{Perceptual Loss}  $\downarrow$ && 0.67 & 0.69 & 0.63 & \textbf{0.59} & 0.60 \\
    \textbf{E$_{warp}$} $\downarrow$ && 73.16 & 69.53 & - & \textbf{49.10} & 56.95 \\
  \bottomrule
\end{tabular*}
\end{table}

\vspace{0.1in}
\noindent \textbf{Perceptual Loss} \textit{Perceptual loss} is usually used in face inpainting approaches~\cite{nazeri2019edgeconnect, yang2019lafin} to measure the similarity between the restored faces and corresponding complete faces. Inspired by this, we use the $relu1\_1$, $relu2\_1$, $relu3\_1$, $relu4\_1 $ and $relu5\_1$ of the pretrained VGG-19 network on ImageNet~\cite{deng2009imagenet} to calculate the perceptual loss between the feature maps of forged faces and that of corresponding real faces. We use the dlib~\cite{king2009dlib} to crop the facial regions. We compare the perceptual loss of different datasets in Table~\ref{table2}. Although the perceptual loss of SR-DF dataset is slightly higher than that of Celeb-DF, it is lower than other datasets by a large margin.

\begin{table*}[t]
\begin{minipage}[t]{0.55\linewidth}
  \caption{Quantitative frame-level detection results on FaceForensics++ dataset under all quality settings. The best results are marked in bold.}
  \label{tab:ff++}
  \vspace{-0.1in}
  \setlength{\tabcolsep}{0pt} 
    \begin{tabular*}{\linewidth}{@{\extracolsep{\fill}}lccccccccc@{}}
    \toprule
    \multirow{2}{*}{\textbf{Methods}} && \multicolumn{2}{c}{\textbf{LQ}} && \multicolumn{2}{c}{\textbf{HQ}} && \multicolumn{2}{c}{\textbf{RAW}}  \\
    ~ && \textbf{ACC} & \textbf{AUC} && \textbf{ACC} & \textbf{AUC} && \textbf{ACC} & \textbf{AUC} \\
    \cmidrule{1-1} \cmidrule{3-4} \cmidrule{6-7} \cmidrule{9-10}
    Steg.Features~\cite{fridrich2012rich} && 55.98 & - && 70.97 & - && 97.63 & - \\
    LD-CNN~\cite{cozzolino2017recasting} && 58.69 & - && 78.45 & - && 98.57 & - \\
    MesoNet~\cite{afchar2018mesonet} && 70.47  & - && 83.10 & - && 95.23 & - \\
    Face X-ray~\cite{li2020face} && - & 61.60 && - & 87.40 && -  & - \\
    F$^{3}$-Net~\cite{qian2020thinking} && 90.43 & 93.30 && 97.52 & 98.10 && \textbf{99.95} & 99.80 \\
    MaDD~\cite{zhao2021multi} && 88.69 & 90.40 && 97.60 & 99.29 && - & - \\
    \system (Xception) && 91.07 & 94.25  && 97.28 & 99.05  && 98.79 & 99.24 \\
    \system (Ours) && \textbf{92.89} & \textbf{95.31} && \textbf{97.93} & \textbf{99.51} && 99.50 & \textbf{99.92} \\
    \bottomrule
\end{tabular*}
\end{minipage}
\hfill
\begin{minipage}[t]{0.4\linewidth}
    \caption{Frame-level AUC scores (\%) of various Deepfake detection methods on Celeb-DF and SR-DF dataset.}
  \label{celebdf_srdf}
  \vspace{-0.1in}
  \renewcommand{\arraystretch}{1.1}
  \setlength{\tabcolsep}{0pt} 
    \begin{tabular*}{\linewidth}{@{\extracolsep{\fill}}lccc@{}}
    \toprule
    \textbf{Methods}  && Celeb-DF~\cite{li2020celeb} & SR-DF  \\
    \cmidrule{1-1} \cmidrule{3-4} 
    Xception~\cite{rossler2019faceforensics++}&& 97.6  &  88.2  \\
    Multi-task~\cite{nguyen2019multi} && 90.5 & 85.7 \\
    Capsule~\cite{nguyen2019use} && 93.2 & 81.5 \\
    DSW-FPA~\cite{li2018exposing}  && 94.8 & 86.6 \\
    DCViT~\cite{wodajo2021deepfake} && 97.2 & 87.9 \\
    Ours && 99.8  & 91.2 \\
    \textbf{Avg} && 95.5 & 86.7 \\
    \bottomrule
\end{tabular*}
\end{minipage} 
\end{table*}

\vspace{0.1in}
\noindent \textbf{Ewarp} The warping error \textbf{$E_{warp}$} is used by~\cite{chen2017coherent, huang2017real, lei2020blind} to measure the temporal inconsistency for video style transfer. We use it to compute the \textbf{$E_{warp}$} of consecutive forged frames in different datasets to quantitatively measure the short-term consistency. Following~\cite{lei2020blind}, we use the method in~\cite{ruder2016artistic} to calculate occlusion map and PWC-Net~\cite{sun2018pwc} to obtain optical flow. $E_{warp}$ of different Deepfake datasets are shown in Table~\ref{table2} for comparison.

\begin{figure}[!ht]
  \centering
  \includegraphics[width=\linewidth]{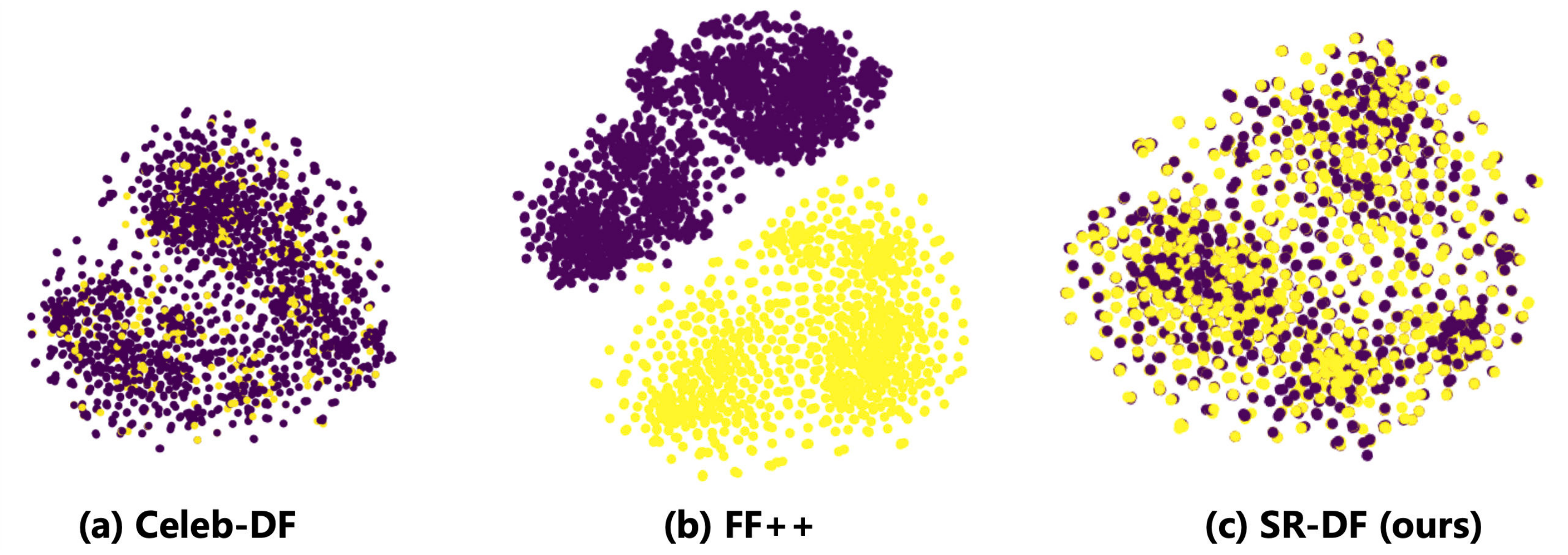}
  \vspace{-0.1in}
  \caption{A feature perspective comparison of Celeb-DF, FF++ dataset (RAW) and SR-DF dataset. We use an ImageNet-pretrained ResNet-18 network to extract features and t-SNE~\cite{van2008visualizing} for dimension reduction. Note that we only select one frame in each video for visualization.}
  \label{figure5}
\end{figure}

\vspace{0.1in}
\noindent \textbf{Feature Space Distribution} As can be seen from the above, Celeb-DF dataset~\cite{li2020celeb} has a decent performance in visual quality. However, they only used one face swapping method to generate all the forged images, which results in limited diversity of data distribution. We illustrate this by visualizing the feature space of Celeb-DF~\cite{li2020celeb}, FF++ dataset~\cite{rossler2019faceforensics++}, and SR-DF in Figure~\ref{figure5}. We can see the data distribution of the Celeb-DF dataset is more concentrated, while the real and forged images of FF++ dataset can be easily separated in the feature space. On the other hand, the data in SR-DF dataset are more scattered in the 2D space.

\section{EXPERIMENTS}
\subsection{Experimental Settings}
\noindent \textbf{Datasets} We conduct experiments on FaceForensics++ (FF++)~\cite{rossler2019faceforensics++}, Celeb-DF~\cite{li2020celeb}, and the proposed SR-DF dataset, and ForgeryNet~\cite{he2021forgerynet}. FF++ consists of 1,000 original videos with real faces, in which 720 videos are used for training, 140 videos are reserved for validation and 140 videos for testing. Each video is manipulated by four Deepfake methods, \ie, Deepfakes~\cite{deepfakes}, FaceSwap~\cite{faceswap}, Face2Face~\cite{Thies_2016_CVPR}, and NeuralTextures~\cite{thies2019deferred}. Different degrees of compression are implemented on both real and forged images to produce high-quality (HQ) version and low-quality (LQ) version of FF++, respectively. Celeb-DF consists of 890 real videos and 5,639 Deepfake videos, in which 6,011 videos are used for training and 518 videos are for testing. ForgeryNet dataset is constructed by 15 manipulation approaches and 36 kinds of distortions for post-processing. It contains 99,630 real videos and 121,617 fake videos. We follow ForgeryNet~\cite{he2021forgerynet} to train our model on the training set, and evaluate on the validation set. For SR-DF, we build on the 1,000 original videos in FF++, and generate 4,000 forged videos using four state-of-the-art subject-agnostic Deepfake generation techniques (see details above). We use the same training, validation and test set partitioning as FF++.

When training on FF++ dataset and SR-DF dataset, following~\cite{qian2020thinking, zhao2021multi}, we augment the real images four times by repeated sampling to balance the number of real and fake samples.  For FF++, we sample 270 frames from each video, following the setting in~\cite{rossler2019faceforensics++, qian2020thinking}. 

\vspace{0.02in}
\noindent \textbf{Evaluation Metrics} We apply the Accuracy score (Acc) and Area Under the RoC Curve (AUC) as our evaluation metrics, which are commonly used in Deepfake detection tasks~\cite{nguyen2019use, rossler2019faceforensics++, li2020face, qian2020thinking, zhao2021multi}.

\vspace{0.02in}
\noindent \textbf{Implementation Details} We use RetinaFace~\cite{deng2020retinaface} to crop the face regions (detected boxes enlarged
1.3$\times$) as inputs with a size of 320 $\times$ 320.  The patch sizes in Sec.4.2.1 are set to (80 $\times$ 80), (40 $\times$ 40), (20 $\times$ 20), and (10 $\times$ 10). For backbone network, we use Efficient-b4~\cite{tan2019efficientnet} pretrained on ImageNet~\cite{deng2009imagenet}. We use Adam for optimization with a learning rate of 0.0001. The learning rate is decayed 10 times every 40 steps. We set the batch size to 24, and train the complete network for 90 epochs. We will release code upon publication.

\begin{table*}[t]
\begin{minipage}[t]{0.4\linewidth}
  \caption{Quantitative comparison of various Deepfake detection methods on ForgeryNet dataset. }
  \label{forgery}
  \vspace{-0.1in}
  \renewcommand{\arraystretch}{1.2}
  \setlength{\tabcolsep}{0pt} 
    \begin{tabular*}{\linewidth}{@{\extracolsep{\fill}}lcccc@{}}
    \toprule
    \textbf{Methods} && \textbf{2 way} & \textbf{3 way} & \textbf{16 way} \\
    
    \cmidrule{1-1} \cmidrule{3-5}
   MobileNetV3~\cite{howard2019searching} && 76.24  & - & - \\
   Xception~\cite{chollet2017xception} && 80.78 & 73.00 & 58.81  \\
   F$^{3}$-Net~\cite{qian2020thinking} && 80.86 & 74.45 & 59.82  \\
   GramNet~\cite{liu2020global} &&  80.89 & 73.30 & 56.77  \\
   SNRFilters-Xception~\cite{chen2017jpeg} &&  81.09  & - & - \\
   Ours && \textbf{82.52} & \textbf{75.12} & \textbf{69.12} \\
    \bottomrule
\end{tabular*}
\end{minipage}
\hfill
\begin{minipage}[t]{0.55\linewidth}
    \caption{Quantitative video-level detection results on FF++ dataset and SR-DF dataset. \system$_{mean}$ denotes averaging the extracted features obtained by \system for all frames as the video-level representation, while \system$_{vtf}$ denotes using VTF Block for temporal fusion.}
  \label{video}
  \vspace{-0.1in}
  \setlength{\tabcolsep}{0pt} 
    \begin{tabular*}{\linewidth}{@{\extracolsep{\fill}}lcccccccc@{}}
    \toprule
    \textbf{Method} && FF++ (RAW) && FF++ (HQ) && FF++ (LQ) && SR-DF \\
    \cmidrule{1-1} \cmidrule{3-3} \cmidrule{5-5} \cmidrule{7-7} \cmidrule{9-9} 
    P3D~\cite{qiu2017learning} && 80.9 && 75.23 && 67.05 && 65.97 \\
    R3D~\cite{xu2017r} && 96.15 && 95.00 && 87.72 && 73.24 \\
    I3D~\cite{carreira2017quo} && 98.23 && 96.70 && 93.18 && 80.11  \\ 
    \system$_{mean}$ && 99.06 && 98.23 && 93.95 && 82.27  \\
    ST-M2TR && \textbf{99.87} && \textbf{99.42} && \textbf{95.31} && \textbf{85.32} \\
  \bottomrule
\end{tabular*}
\end{minipage} 
\end{table*}

\begin{table*}[t]
  \caption{AUC scores (\%) for cross-dataset evaluation on FF++, Celeb-DF, and SR-DF datasets. Note that some methods have not made their code public, so we directly use the data reported in their paper. ``$-$'' denotes the results are unavailable.}
  \vspace{-0.1in}
  \label{cross}
  \resizebox{\linewidth}{!}{\begin{tabular}{ccccccccccccc}
    \toprule
    \textbf{Training Set} && \textbf{Testing Set} && Xception~\cite{rossler2019faceforensics++} & Multi-task~\cite{nguyen2019multi} & Capsule~\cite{nguyen2019use} & DSW-FPA~\cite{li2018exposing} & Two-Branch~\cite{masi2020two}  & F3-Net ~\cite{qian2020thinking}  & MaDD~\cite{zhao2021multi}  & DCViT~\cite{wodajo2021deepfake} & Ours \\
    \cmidrule{1-1} \cmidrule{3-3} \cmidrule{5-13}
    \multirow{2}{*}{FF++} && FF++ && 99.7 & 76.3 & 96.6 & 93.0 & 98.7 & 98.1 & 99.3 & 98.3 & \textbf{ 99.5} \\
    
    ~ && Celeb-DF && 48.2 & 54.3 & 57.5 & 64.6 & \textbf{73.4}	& 65.2 & 67.4 & 60.8 & 68.2 \\
    
    ~ && SR-DF && 37.9 & 38.7 & 41.3 & 44.0	& - & - & -	& 57.8 &  \textbf{63.7} \\
    \cmidrule{1-1} \cmidrule{3-3} \cmidrule{5-13}
    \multirow{2}{*}{SR-DF} && SR-DF && 88.2 & 85.7 & 81.5 & 86.6 & - & - & - & 87.9 & \textbf{91.2} \\  
    
    ~ && FF++ && 63.2 & 58.9 & 60.6 & 69.1 & - & - & - & 62.6 & \textbf{79.7} \\
    
    ~ && Celeb-DF && 59.4 & 51.7 & 52.1 & 62.9 & - & - & -	& 63.7 & \textbf{82.1} \\
    \bottomrule
\end{tabular}}
\end{table*}

\subsection{Evaluation on FaceForensics++}
FF++~\cite{rossler2019faceforensics++} is a widely used dataset in various Deepfake detection approaches~\cite{fridrich2012rich, cozzolino2017recasting, afchar2018mesonet, li2020face, qian2020thinking, zhao2021multi}. We compare \system with top-notch methods on it, including: Steg. Features~\cite{fridrich2012rich}, LD-CNN~\cite{cozzolino2017recasting}, MesoNet~\cite{afchar2018mesonet}, Face X-ray~\cite{li2020face}, F$^{3}$-Net~\cite{qian2020thinking}, and MaDD~\cite{zhao2021multi}.

We test the frame-level detection performance on RAW, HQ, and LQ of FF++, respectively, and report the AUC scores (\%) in Table \ref{tab:ff++}. We can see that our method achieves state-of-the-art performance on all versions (\ie, LQ, HQ, and RAW), which suggests the effectiveness of our approach in detecting Deepfakes of different visual qualities. We also evaluate \system using different backbones, and the performances verifies that our framework is not restricted by the backbone
networks. Comparing across different versions of the FF++ dataset, we see that 
while most approaches achieve high performance on the high-quality version of FF++, we observe a significant performance degradation on FF++ (LQ) where the forged images are compressed. This could be remedied by leveraging frequency information. While both F3-Net and \system use frequency features, \system achieves an accuracy of 92.35\% in the LQ setting, outperforming the F3-Net approach by 1.92\%.

\subsection{Evaluation on Celeb-DF and SR-DF}
In this section, we conduct experiments to evaluate the detection accuracy of our \system on Celeb-DF~\cite{li2020celeb} and SR-DF dataset at frame-level, respectively. Note that we do not report the quantitative results of certain state-of-the-art Deepfake detection methods including~\cite{li2020face, qian2020thinking, zhao2021multi} because the code and models are not publicly available. The results are reported in Table~\ref{celebdf_srdf}. We observe that our \system achieves 99.9\% and 90.5\% on Celeb-DF and SR-DF, respectively, which demonstrate that our method outperforms all the other Deepfake detection methods over different datasets. This suggests that our approach is indeed effective for Deepfake detection across different datasets.

In addition, the quality of different Deepfake datasets can be evaluated by comparing the detection accuracy of the same detection method on different datasets. Given that Celeb-DF~\cite{li2020celeb} contains high-quality samples  (as discussed in 4.2, Celeb-DF achieves the best results on \textit{Mask-SSIM}, \textit{Perceptual loss} and \textit{Ewarp} metrics in the available Deepfake dataset.), we calculate the average frame-level AUC scores of all compared detection methods on Celeb-DF dataset and SR-DF, and report them in the last row of Table~\ref{celebdf_srdf}. The overall performance on SR-DF is 9.2\% lower than that of Celeb-DF, which demonstrates that SR-DF is more challenging. 

\subsection{Evaluation on ForgeryNet}
ForgeryNet~\cite{he2021forgerynet} is the most recently released largest scale deepfake detection dataset, which provides three types of forgery labels, \ie, two-way (real / fake), three-way (real / fake with identity-replaced
forgery approaches / fake with identity-remained forgery
approaches), and n-way (n = 16, real and 15 respective
forgery approaches) labels. Using the rich annotations, we conduct two-/three-/n-way classification experiments. The comparison results in Table~\ref{forgery} demonstrate that \system outperforms state-of-the-art methods on classification tasks of different granularity. In particular, the most remarkable performance gain \ie, 15.5\% compared with F$^{3}$-Net, is achieved on the most challenging 16-way classification, which is benefited from the capability of our method to identify the multi-scale forgery features.

\subsection{From Frames to Videos}
Existing methods on Deepfake detection mainly perform evaluation based on frames extracted from videos, albeit videos are provided. However, in real-world scenarios, most Deepfake data circulating on the Internet are fake videos, therefore, we also conduct experiments to evaluate our \system on video-level Deepfake detection. The most significant difference between videos and images is the additional temporal information between frame sequences. We demonstrate that \system can be easily extended for video modeling by adding a temporal transformer to combine frame-level features generated by \system. We refer to such an extension as spatial-temporal M2TR (ST-M2TR).

In particular, we sampled 16 frames at intervals from one video, and directly use the model trained at the frame-level to extract features of different frames. These features are then input to a transformer block (it has 4 stacked encoders, each with 8 attention heads, and an MLP head that has two fc layers) to obtain video-level predictions. We report the AUC scores (\%) and compare with (1) P3D~\cite{qiu2017learning}, which simplifies 3D convolutions with 2D filters on spatial dimension and 1D temporal connections; (2)R3D~\cite{xu2017r}, which encodes the video sequences using a 3D fully convolutional networks and then generates candidate temporal fragments for  classification; (3)I3D~\cite{carreira2017quo}, which expands 2D CNNs with an additional temporal dimension to introduce a two-stream inflated 3D convolutional network; (4) M2TR$_{mean}$, which averages the features of different frames by \system for video-level prediction. Note that (1) and (3) are designed for video action recognition, while (2) is for temporal activity detection, and we modify them for video-level Deepfake detection. The results are summarized in Table~\ref{video}. We can see that our method achieves the best performance on FF++, ForgeryNet, and SR-DF.

\subsection{Generalization Ability}
The generalization ability is at the core of Deepfake detection. We evaluate the generalization of our \system by separately training on FF++ (HQ) and SR-DF dataset, and test on other datasets. We follow~\cite{zhao2021multi} to sample 30 frames for each video and calculate the frame-level AUC scores. The comparison results are shown in Table~\ref{cross}. Note that for the Deepfake detection models that are not publicly available, we only use the results reported in their paper. The results in Table~\ref{cross} demonstrate that our method achieves better generalization than most existing methods.

\begin{table*}[t]
\begin{minipage}[t]{0.55\linewidth}
\centering
  \caption{Ablation results on FF++ (HQ) and FF++ (LQ) with and without Multi-scale Transformer and CMF.}
  \vspace{-0.1in}
  \label{mt_st}
  \setlength{\tabcolsep}{0pt} 
    \begin{tabular*}{\linewidth}{@{\extracolsep{\fill}}lcccccc@{}}
    \toprule
    \multirow{2}{*}{\textbf{Method}} && \multicolumn{2}{c}{LQ} && \multicolumn{2}{c}{HQ} \\
    ~ && ACC (\%) & AUC (\%) && ACC (\%) & AUC (\%) \\
     \cmidrule{1-1} \cmidrule{3-4} \cmidrule{6-7}
    w/o MT && 87.19 & 90.05 && 94.88 & 96.94 \\
    w/o FF && 89.33 & 90.48 && 95.89  & 97.94 \\ 
    w/o CMF && 90.70 & 92.37 && 96.78 & 98.19 \\
    Ours && \textbf{92.89} & \textbf{95.31} && \textbf{97.93} & \textbf{99.51}\\
  \bottomrule
\end{tabular*}
\end{minipage}
\hfill
\begin{minipage}[t]{0.4\linewidth}
\centering
 \caption{Ablation results on FF++ (HQ) using multi-scale Transformer (MT) or single-scale transformer.}
  \vspace{-0.1in}
  \label{patch-size}
  \setlength{\tabcolsep}{0pt} 
    \begin{tabular*}{\linewidth}{@{\extracolsep{\fill}}lccc@{}}
    \toprule
    \textbf{Patch size} && ACC (\%) & AUC (\%) \\
     \cmidrule{1-1} \cmidrule{3-4} 
     $40\times40$ && 93.58 & 95.81 \\
     $20\times20$ && 96.65 & 97.53 \\
     $10\times10$  && 97.19 & 98.83 \\
     Ours && \textbf{97.93} & \textbf{99.51} \\
  \bottomrule
\end{tabular*}
\end{minipage}
\end{table*}

\subsection{Ablation Study}
\noindent \textbf{Effectiveness of Different Components} The Multi-scale Transformer (MT) of our method is designed to capture local inconsistency between patches of different sizes, while the Frequency Filter (FF) is utilized to capture the subtle forgery traces in frequency domain. To evaluate the effectiveness of MT and FF, we remove them separately from \system and demonstrate the performance degradation on FF++. We also replace the Cross Modality Fusion (CMF) blocks with naive concatenation operations to verify the usefulness of CMF for feature fusion. 

The quantitative results are listed in Table~\ref{mt_st}, which validates that the use of MT, FF and CMF can effectively improve the detection performance of our model. In particular, the proposed frequency filter brings a remarkable improvement to our method under the low-quality (LQ) setting, \ie, about 1.7\% performance gain on AUC score, which is mainly benefited from the complementary information from the frequency modality.

\vspace{0.1in}
\noindent \textbf{Effectiveness of the Multi-scale Design} To verify the effectiveness of using multi-scale patches in different heads in our multi-scale transformer, we replace MT with several single-scale transformers with different patch sizes, and conduct experiments on FF++ (HQ). The results in Table~\ref{patch-size} demonstrate that our full model achieves the best performance with MT, \ie, 3.9\%, 1.9\%, and 0.7\% higher than $40\times40$, $20\times20$ and $10\times10$ single-scale transformer on AUC score. This confirms the use of a multi-scale transformer is indeed effective.

\vspace{0.1in}
\noindent \textbf{Effectiveness of the Contrastive Loss} 
To illustrate the contribution of contrastive loss in improving the generalization ability of our method, we also conduct experiments to train \system without its supervision and evaluate the cross-dataset detection accuracy. The comparison results are reported in Table~\ref{constrative}. We can observe that 1) When training on FF++ without the contrastive loss, the accuracy decreases by 3.8\% and 5.2\% in Celeb-DF and SR-DF, respectively. 2) When training on SR-DF dataset without the contrastive loss, the accuracy decreases by 5.8\% and 3.0\%, respectively. 

\begin{table}[!ht]
\centering
  \caption{AUC (\%) for cross-dataset evaluation on FF++ (HQ), Celeb-DF, and SR-DF with (denoted as \system) and without (denoted as \system$_{ncl}$) the supervision of contrastive loss.} 
  \vspace{-0.1in}
  \label{constrative}
  \setlength{\tabcolsep}{0pt} 
    \begin{tabular*}{\linewidth}{@{\extracolsep{\fill}}cccccc@{}}
    \toprule
    \textbf{Training Set} && \textbf{Testing Set} && \system$_{ncl}$ & \system \\
    \cmidrule{1-1} \cmidrule{3-3} \cmidrule{5-6}
    \multirow{2}{*}{FF++} && Celeb-DF && 65.6  & 68.2 \\
    
    ~ && SR-DF && 60.4 &  63.7 \\
    \cmidrule{1-1} \cmidrule{3-3} \cmidrule{5-6}
    \multirow{2}{*}{SR-DF} && FF++ && 75.1  & 79.7 \\
    
    ~ && Celeb-DF && 79.6 & 82.1 \\
    \bottomrule
\end{tabular*}
\end{table}

\section{Conclusion}
In this paper, we presented a two-stream network Multi-modal Multi-scale Transformer (\system) for Deepfake detection, which uses multi-scale transformers to capture subtle local inconsistency at multiple scales and frequency filters to improve the robustness against image compression. Forgery feature from two streams are adaptively fused through cross modality fusion blocks. Besides, we introduced a challenging dataset SR-DF that are generated with several state-of-the-art face swapping and facial reenactment methods. We also built the most comprehensive evaluation system to  quantitatively verify that the SR-DF dataset is better than existing datasets in terms of visual quality and data diversity. Extensive experiments on different datasets demonstrate the effectiveness of the proposed method. 

\vspace{0.1in}
\noindent \textbf{Acknowledgement}
This work was supported by National Natural Science Foundation of Project (62072116). Y.-G. Jiang was sponsored in part by ``Shuguang Program'' supported by Shanghai Education Development Foundation and Shanghai Municipal Education Commission (20SG01).

\bibliographystyle{ACM-Reference-Format}
\bibliography{main}


\begin{thebibliography}{74}


\ifx \showCODEN    \undefined \def \showCODEN     #1{\unskip}     \fi
\ifx \showDOI      \undefined \def \showDOI       #1{#1}\fi
\ifx \showISBNx    \undefined \def \showISBNx     #1{\unskip}     \fi
\ifx \showISBNxiii \undefined \def \showISBNxiii  #1{\unskip}     \fi
\ifx \showISSN     \undefined \def \showISSN      #1{\unskip}     \fi
\ifx \showLCCN     \undefined \def \showLCCN      #1{\unskip}     \fi
\ifx \shownote     \undefined \def \shownote      #1{#1}          \fi
\ifx \showarticletitle \undefined \def \showarticletitle #1{#1}   \fi
\ifx \showURL      \undefined \def \showURL       {\relax}        \fi
\providecommand\bibfield[2]{#2}
\providecommand\bibinfo[2]{#2}
\providecommand\natexlab[1]{#1}
\providecommand\showeprint[2][]{arXiv:#2}

\bibitem[\protect\citeauthoryear{Afchar, Nozick, Yamagishi, and Echizen}{Afchar
  et~al\mbox{.}}{2018}]%
        {afchar2018mesonet}
\bibfield{author}{\bibinfo{person}{Darius Afchar}, \bibinfo{person}{Vincent
  Nozick}, \bibinfo{person}{Junichi Yamagishi}, {and} \bibinfo{person}{Isao
  Echizen}.} \bibinfo{year}{2018}\natexlab{}.
\newblock \showarticletitle{Mesonet: a compact facial video forgery detection
  network}. In \bibinfo{booktitle}{\emph{WIFS}}.
\newblock


\bibitem[\protect\citeauthoryear{Carion, Massa, Synnaeve, Usunier, Kirillov,
  and Zagoruyko}{Carion et~al\mbox{.}}{2020}]%
        {carion2020end}
\bibfield{author}{\bibinfo{person}{Nicolas Carion}, \bibinfo{person}{Francisco
  Massa}, \bibinfo{person}{Gabriel Synnaeve}, \bibinfo{person}{Nicolas
  Usunier}, \bibinfo{person}{Alexander Kirillov}, {and} \bibinfo{person}{Sergey
  Zagoruyko}.} \bibinfo{year}{2020}\natexlab{}.
\newblock \showarticletitle{End-to-end object detection with transformers}. In
  \bibinfo{booktitle}{\emph{ECCV}}.
\newblock


\bibitem[\protect\citeauthoryear{Carreira and Zisserman}{Carreira and
  Zisserman}{2017}]%
        {carreira2017quo}
\bibfield{author}{\bibinfo{person}{Joao Carreira} {and} \bibinfo{person}{Andrew
  Zisserman}.} \bibinfo{year}{2017}\natexlab{}.
\newblock \showarticletitle{Quo vadis, action recognition? a new model and the
  kinetics dataset}. In \bibinfo{booktitle}{\emph{CVPR}}.
\newblock


\bibitem[\protect\citeauthoryear{Chen, Liao, Yuan, Yu, and Hua}{Chen
  et~al\mbox{.}}{2017a}]%
        {chen2017coherent}
\bibfield{author}{\bibinfo{person}{Dongdong Chen}, \bibinfo{person}{Jing Liao},
  \bibinfo{person}{Lu Yuan}, \bibinfo{person}{Nenghai Yu}, {and}
  \bibinfo{person}{Gang Hua}.} \bibinfo{year}{2017}\natexlab{a}.
\newblock \showarticletitle{Coherent online video style transfer}. In
  \bibinfo{booktitle}{\emph{ICCV}}.
\newblock


\bibitem[\protect\citeauthoryear{Chen, Sedighi, Boroumand, and Fridrich}{Chen
  et~al\mbox{.}}{2017b}]%
        {chen2017jpeg}
\bibfield{author}{\bibinfo{person}{Mo Chen}, \bibinfo{person}{Vahid Sedighi},
  \bibinfo{person}{Mehdi Boroumand}, {and} \bibinfo{person}{Jessica Fridrich}.}
  \bibinfo{year}{2017}\natexlab{b}.
\newblock \showarticletitle{JPEG-phase-aware convolutional neural network for
  steganalysis of JPEG images}. In \bibinfo{booktitle}{\emph{IHMSW}}.
\newblock


\bibitem[\protect\citeauthoryear{Chen, Yao, Chen, Ding, Li, and Ji}{Chen
  et~al\mbox{.}}{2021}]%
        {chen2021local}
\bibfield{author}{\bibinfo{person}{Shen Chen}, \bibinfo{person}{Taiping Yao},
  \bibinfo{person}{Yang Chen}, \bibinfo{person}{Shouhong Ding},
  \bibinfo{person}{Jilin Li}, {and} \bibinfo{person}{Rongrong Ji}.}
  \bibinfo{year}{2021}\natexlab{}.
\newblock \showarticletitle{Local Relation Learning for Face Forgery
  Detection}. In \bibinfo{booktitle}{\emph{AAAI}}.
\newblock


\bibitem[\protect\citeauthoryear{Chollet}{Chollet}{2017}]%
        {chollet2017xception}
\bibfield{author}{\bibinfo{person}{Fran{\c{c}}ois Chollet}.}
  \bibinfo{year}{2017}\natexlab{}.
\newblock \showarticletitle{Xception: Deep learning with depthwise separable
  convolutions}. In \bibinfo{booktitle}{\emph{CVPR}}.
\newblock


\bibitem[\protect\citeauthoryear{Cong, Zhang, Niu, Liu, Ling, Li, and
  Zhang}{Cong et~al\mbox{.}}{2020}]%
        {cong2020dovenet}
\bibfield{author}{\bibinfo{person}{Wenyan Cong}, \bibinfo{person}{Jianfu
  Zhang}, \bibinfo{person}{Li Niu}, \bibinfo{person}{Liu Liu},
  \bibinfo{person}{Zhixin Ling}, \bibinfo{person}{Weiyuan Li}, {and}
  \bibinfo{person}{Liqing Zhang}.} \bibinfo{year}{2020}\natexlab{}.
\newblock \showarticletitle{DoveNet: Deep Image Harmonization via Domain
  Verification}. In \bibinfo{booktitle}{\emph{CVPR}}.
\newblock


\bibitem[\protect\citeauthoryear{Cozzolino, Poggi, and Verdoliva}{Cozzolino
  et~al\mbox{.}}{2017}]%
        {cozzolino2017recasting}
\bibfield{author}{\bibinfo{person}{Davide Cozzolino}, \bibinfo{person}{Giovanni
  Poggi}, {and} \bibinfo{person}{Luisa Verdoliva}.}
  \bibinfo{year}{2017}\natexlab{}.
\newblock \showarticletitle{Recasting residual-based local descriptors as
  convolutional neural networks: an application to image forgery detection}. In
  \bibinfo{booktitle}{\emph{Workshop on IH\&MMSec}}.
\newblock


\bibitem[\protect\citeauthoryear{Dataset}{Dataset}{2019}]%
        {dfd}
\bibfield{author}{\bibinfo{person}{DeepFake~Detection Dataset}.}
  \bibinfo{year}{2019}\natexlab{}.
\newblock
  \bibinfo{howpublished}{\url{https://ai.googleblog.com/2019/09/contributing-data-to-deepfake-detection.html}}.
\newblock


\bibitem[\protect\citeauthoryear{Deepfakes}{Deepfakes}{2018}]%
        {deepfakes}
\bibfield{author}{\bibinfo{person}{Deepfakes}.}
  \bibinfo{year}{2018}\natexlab{}.
\newblock \bibinfo{title}{github}.
\newblock \bibinfo{howpublished}{\url{https://github.com/deepfakes/faceswap}}.
\newblock


\bibitem[\protect\citeauthoryear{Deng, Dong, Socher, Li, Li, and Fei-Fei}{Deng
  et~al\mbox{.}}{2009}]%
        {deng2009imagenet}
\bibfield{author}{\bibinfo{person}{Jia Deng}, \bibinfo{person}{Wei Dong},
  \bibinfo{person}{Richard Socher}, \bibinfo{person}{Li-Jia Li},
  \bibinfo{person}{Kai Li}, {and} \bibinfo{person}{Li Fei-Fei}.}
  \bibinfo{year}{2009}\natexlab{}.
\newblock \showarticletitle{Imagenet: A large-scale hierarchical image
  database}. In \bibinfo{booktitle}{\emph{CVPR}}.
\newblock


\bibitem[\protect\citeauthoryear{Deng, Guo, Ververas, Kotsia, and
  Zafeiriou}{Deng et~al\mbox{.}}{2020}]%
        {deng2020retinaface}
\bibfield{author}{\bibinfo{person}{Jiankang Deng}, \bibinfo{person}{Jia Guo},
  \bibinfo{person}{Evangelos Ververas}, \bibinfo{person}{Irene Kotsia}, {and}
  \bibinfo{person}{Stefanos Zafeiriou}.} \bibinfo{year}{2020}\natexlab{}.
\newblock \showarticletitle{Retinaface: Single-shot multi-level face
  localisation in the wild}. In \bibinfo{booktitle}{\emph{CVPR}}.
\newblock


\bibitem[\protect\citeauthoryear{Devlin, Chang, Lee, and Toutanova}{Devlin
  et~al\mbox{.}}{2019}]%
        {devlin2018bert}
\bibfield{author}{\bibinfo{person}{Jacob Devlin}, \bibinfo{person}{Ming-Wei
  Chang}, \bibinfo{person}{Kenton Lee}, {and} \bibinfo{person}{Kristina
  Toutanova}.} \bibinfo{year}{2019}\natexlab{}.
\newblock \showarticletitle{BERT: Pre-training of Deep Bidirectional
  Transformers for Language Understanding}. In
  \bibinfo{booktitle}{\emph{NAACL-HLT}}.
\newblock


\bibitem[\protect\citeauthoryear{Dolhansky, Howes, Pflaum, Baram, and
  Ferrer}{Dolhansky et~al\mbox{.}}{2020}]%
        {dolhansky2020deepfake}
\bibfield{author}{\bibinfo{person}{Brian Dolhansky}, \bibinfo{person}{Russ
  Howes}, \bibinfo{person}{Ben Pflaum}, \bibinfo{person}{Nicole Baram}, {and}
  \bibinfo{person}{Cristian~Canton Ferrer}.} \bibinfo{year}{2020}\natexlab{}.
\newblock \showarticletitle{The deepfake detection challenge (dfdc) dataset}.
\newblock \bibinfo{journal}{\emph{arXiv preprint arXiv:2006.07397}}
  (\bibinfo{year}{2020}).
\newblock


\bibitem[\protect\citeauthoryear{Dosovitskiy, Beyer, Kolesnikov, Weissenborn,
  Zhai, Unterthiner, Dehghani, Minderer, Heigold, Gelly, Uszkoreit, and
  Houlsby}{Dosovitskiy et~al\mbox{.}}{2021}]%
        {dosovitskiy2020image}
\bibfield{author}{\bibinfo{person}{Alexey Dosovitskiy}, \bibinfo{person}{Lucas
  Beyer}, \bibinfo{person}{Alexander Kolesnikov}, \bibinfo{person}{Dirk
  Weissenborn}, \bibinfo{person}{Xiaohua Zhai}, \bibinfo{person}{Thomas
  Unterthiner}, \bibinfo{person}{Mostafa Dehghani}, \bibinfo{person}{Matthias
  Minderer}, \bibinfo{person}{Georg Heigold}, \bibinfo{person}{Sylvain Gelly},
  \bibinfo{person}{Jakob Uszkoreit}, {and} \bibinfo{person}{Neil Houlsby}.}
  \bibinfo{year}{2021}\natexlab{}.
\newblock \showarticletitle{An Image is Worth 16x16 Words: Transformers for
  Image Recognition at Scale}. In \bibinfo{booktitle}{\emph{ICLR}}.
\newblock


\bibitem[\protect\citeauthoryear{Durall, Keuper, Pfreundt, and Keuper}{Durall
  et~al\mbox{.}}{2019}]%
        {durall2019unmasking}
\bibfield{author}{\bibinfo{person}{Ricard Durall}, \bibinfo{person}{Margret
  Keuper}, \bibinfo{person}{Franz-Josef Pfreundt}, {and} \bibinfo{person}{Janis
  Keuper}.} \bibinfo{year}{2019}\natexlab{}.
\newblock \showarticletitle{Unmasking deepfakes with simple features}.
\newblock \bibinfo{journal}{\emph{arXiv preprint arXiv:1911.00686}}
  (\bibinfo{year}{2019}).
\newblock


\bibitem[\protect\citeauthoryear{Face-parsing}{Face-parsing}{2019}]%
        {face-parsing}
\bibfield{author}{\bibinfo{person}{Face-parsing}.}
  \bibinfo{year}{2019}\natexlab{}.
\newblock \bibinfo{title}{github}.
\newblock
  \bibinfo{howpublished}{\url{https://github.com/zllrunning/face-parsing.PyTorch}}.
\newblock


\bibitem[\protect\citeauthoryear{FaceShifter}{FaceShifter}{2020}]%
        {faceshifter}
\bibfield{author}{\bibinfo{person}{FaceShifter}.}
  \bibinfo{year}{2020}\natexlab{}.
\newblock \bibinfo{title}{github}.
\newblock
  \bibinfo{howpublished}{\url{https://github.com/mindslab-ai/faceshifter}}.
\newblock


\bibitem[\protect\citeauthoryear{Faceswap}{Faceswap}{2018}]%
        {faceswap}
\bibfield{author}{\bibinfo{person}{Faceswap}.} \bibinfo{year}{2018}\natexlab{}.
\newblock \bibinfo{title}{github}.
\newblock
  \bibinfo{howpublished}{\url{https://github.com/MarekKowalski/FaceSwap/}}.
\newblock


\bibitem[\protect\citeauthoryear{Fridrich and Kodovsky}{Fridrich and
  Kodovsky}{2012}]%
        {fridrich2012rich}
\bibfield{author}{\bibinfo{person}{Jessica Fridrich} {and} \bibinfo{person}{Jan
  Kodovsky}.} \bibinfo{year}{2012}\natexlab{}.
\newblock \showarticletitle{Rich models for steganalysis of digital images}.
\newblock \bibinfo{journal}{\emph{TIFS}} (\bibinfo{year}{2012}).
\newblock


\bibitem[\protect\citeauthoryear{He, Zhang, Ren, and Sun}{He
  et~al\mbox{.}}{2016}]%
        {he2016deep}
\bibfield{author}{\bibinfo{person}{Kaiming He}, \bibinfo{person}{Xiangyu
  Zhang}, \bibinfo{person}{Shaoqing Ren}, {and} \bibinfo{person}{Jian Sun}.}
  \bibinfo{year}{2016}\natexlab{}.
\newblock \showarticletitle{Deep residual learning for image recognition}. In
  \bibinfo{booktitle}{\emph{CVPR}}.
\newblock


\bibitem[\protect\citeauthoryear{He, Gan, Chen, Zhou, Yin, Song, Sheng, Shao,
  and Liu}{He et~al\mbox{.}}{2021}]%
        {he2021forgerynet}
\bibfield{author}{\bibinfo{person}{Yinan He}, \bibinfo{person}{Bei Gan},
  \bibinfo{person}{Siyu Chen}, \bibinfo{person}{Yichun Zhou},
  \bibinfo{person}{Guojun Yin}, \bibinfo{person}{Luchuan Song},
  \bibinfo{person}{Lu Sheng}, \bibinfo{person}{Jing Shao}, {and}
  \bibinfo{person}{Ziwei Liu}.} \bibinfo{year}{2021}\natexlab{}.
\newblock \showarticletitle{ForgeryNet: A Versatile Benchmark for Comprehensive
  Forgery Analysis}. In \bibinfo{booktitle}{\emph{CVPR}}.
\newblock


\bibitem[\protect\citeauthoryear{Howard, Sandler, Chu, Chen, Chen, Tan, Wang,
  Zhu, Pang, Vasudevan, et~al\mbox{.}}{Howard et~al\mbox{.}}{2019}]%
        {howard2019searching}
\bibfield{author}{\bibinfo{person}{Andrew Howard}, \bibinfo{person}{Mark
  Sandler}, \bibinfo{person}{Grace Chu}, \bibinfo{person}{Liang-Chieh Chen},
  \bibinfo{person}{Bo Chen}, \bibinfo{person}{Mingxing Tan},
  \bibinfo{person}{Weijun Wang}, \bibinfo{person}{Yukun Zhu},
  \bibinfo{person}{Ruoming Pang}, \bibinfo{person}{Vijay Vasudevan},
  {et~al\mbox{.}}} \bibinfo{year}{2019}\natexlab{}.
\newblock \showarticletitle{Searching for mobilenetv3}. In
  \bibinfo{booktitle}{\emph{ICCV}}.
\newblock


\bibitem[\protect\citeauthoryear{Huang, Wang, Luo, Ma, Jiang, Zhu, Li, and
  Liu}{Huang et~al\mbox{.}}{2017}]%
        {huang2017real}
\bibfield{author}{\bibinfo{person}{Haozhi Huang}, \bibinfo{person}{Hao Wang},
  \bibinfo{person}{Wenhan Luo}, \bibinfo{person}{Lin Ma},
  \bibinfo{person}{Wenhao Jiang}, \bibinfo{person}{Xiaolong Zhu},
  \bibinfo{person}{Zhifeng Li}, {and} \bibinfo{person}{Wei Liu}.}
  \bibinfo{year}{2017}\natexlab{}.
\newblock \showarticletitle{Real-time neural style transfer for videos}. In
  \bibinfo{booktitle}{\emph{CVPR}}.
\newblock


\bibitem[\protect\citeauthoryear{Huang, Zhang, and Wang}{Huang
  et~al\mbox{.}}{2020}]%
        {huang2020deep}
\bibfield{author}{\bibinfo{person}{Ying Huang}, \bibinfo{person}{Wenwei Zhang},
  {and} \bibinfo{person}{Jinzhuo Wang}.} \bibinfo{year}{2020}\natexlab{}.
\newblock \showarticletitle{Deep frequent spatial temporal learning for face
  anti-spoofing}.
\newblock \bibinfo{journal}{\emph{arXiv preprint arXiv:2002.03723}}
  (\bibinfo{year}{2020}).
\newblock


\bibitem[\protect\citeauthoryear{Jeon, Bang, and Woo}{Jeon
  et~al\mbox{.}}{2020}]%
        {jeon2020fdftnet}
\bibfield{author}{\bibinfo{person}{Hyeonseong Jeon}, \bibinfo{person}{Youngoh
  Bang}, {and} \bibinfo{person}{Simon~S Woo}.} \bibinfo{year}{2020}\natexlab{}.
\newblock \showarticletitle{FDFtNet: Facing off fake images using fake
  detection fine-tuning network}. In \bibinfo{booktitle}{\emph{ICT Systems
  Security and Privacy Protection}}.
\newblock


\bibitem[\protect\citeauthoryear{Kemelmacher-Shlizerman}{Kemelmacher-Shlizerman}{2016}]%
        {kemelmacher2016transfiguring}
\bibfield{author}{\bibinfo{person}{Ira Kemelmacher-Shlizerman}.}
  \bibinfo{year}{2016}\natexlab{}.
\newblock \showarticletitle{Transfiguring portraits}.
\newblock \bibinfo{journal}{\emph{ACM TOG}} (\bibinfo{year}{2016}).
\newblock


\bibitem[\protect\citeauthoryear{King}{King}{2009}]%
        {king2009dlib}
\bibfield{author}{\bibinfo{person}{Davis~E King}.}
  \bibinfo{year}{2009}\natexlab{}.
\newblock \showarticletitle{Dlib-ml: A machine learning toolkit}.
\newblock \bibinfo{journal}{\emph{JMLR}} (\bibinfo{year}{2009}).
\newblock


\bibitem[\protect\citeauthoryear{Korshunov and Marcel}{Korshunov and
  Marcel}{2018}]%
        {korshunov2018deepfakes}
\bibfield{author}{\bibinfo{person}{Pavel Korshunov} {and}
  \bibinfo{person}{S{\'e}bastien Marcel}.} \bibinfo{year}{2018}\natexlab{}.
\newblock \showarticletitle{Deepfakes: a new threat to face recognition?
  assessment and detection}.
\newblock \bibinfo{journal}{\emph{arXiv preprint arXiv:1812.08685}}
  (\bibinfo{year}{2018}).
\newblock


\bibitem[\protect\citeauthoryear{Koujan, Doukas, Roussos, and Zafeiriou}{Koujan
  et~al\mbox{.}}{2020}]%
        {koujan2020head2head}
\bibfield{author}{\bibinfo{person}{Mohammad~Rami Koujan},
  \bibinfo{person}{Michail~Christos Doukas}, \bibinfo{person}{Anastasios
  Roussos}, {and} \bibinfo{person}{Stefanos Zafeiriou}.}
  \bibinfo{year}{2020}\natexlab{}.
\newblock \showarticletitle{Head2head: Video-based neural head synthesis}.
\newblock \bibinfo{journal}{\emph{arXiv preprint arXiv:2005.10954}}
  (\bibinfo{year}{2020}).
\newblock


\bibitem[\protect\citeauthoryear{Lei, Xing, and Chen}{Lei
  et~al\mbox{.}}{2020}]%
        {lei2020blind}
\bibfield{author}{\bibinfo{person}{Chenyang Lei}, \bibinfo{person}{Yazhou
  Xing}, {and} \bibinfo{person}{Qifeng Chen}.} \bibinfo{year}{2020}\natexlab{}.
\newblock \showarticletitle{Blind Video Temporal Consistency via Deep Video
  Prior}. In \bibinfo{booktitle}{\emph{NIPS}}.
\newblock


\bibitem[\protect\citeauthoryear{Li, Bao, Yang, Chen, and Wen}{Li
  et~al\mbox{.}}{2019}]%
        {li2019faceshifter}
\bibfield{author}{\bibinfo{person}{Lingzhi Li}, \bibinfo{person}{Jianmin Bao},
  \bibinfo{person}{Hao Yang}, \bibinfo{person}{Dong Chen}, {and}
  \bibinfo{person}{Fang Wen}.} \bibinfo{year}{2019}\natexlab{}.
\newblock \showarticletitle{Faceshifter: Towards high fidelity and occlusion
  aware face swapping}.
\newblock \bibinfo{journal}{\emph{arXiv preprint arXiv:1912.13457}}
  (\bibinfo{year}{2019}).
\newblock


\bibitem[\protect\citeauthoryear{Li, Bao, Zhang, Yang, Chen, Wen, and Guo}{Li
  et~al\mbox{.}}{2020a}]%
        {li2020face}
\bibfield{author}{\bibinfo{person}{Lingzhi Li}, \bibinfo{person}{Jianmin Bao},
  \bibinfo{person}{Ting Zhang}, \bibinfo{person}{Hao Yang},
  \bibinfo{person}{Dong Chen}, \bibinfo{person}{Fang Wen}, {and}
  \bibinfo{person}{Baining Guo}.} \bibinfo{year}{2020}\natexlab{a}.
\newblock \showarticletitle{Face x-ray for more general face forgery
  detection}. In \bibinfo{booktitle}{\emph{CVPR}}.
\newblock


\bibitem[\protect\citeauthoryear{Li and Lyu}{Li and Lyu}{2018}]%
        {li2018exposing}
\bibfield{author}{\bibinfo{person}{Yuezun Li} {and} \bibinfo{person}{Siwei
  Lyu}.} \bibinfo{year}{2018}\natexlab{}.
\newblock \showarticletitle{Exposing deepfake videos by detecting face warping
  artifacts}.
\newblock \bibinfo{journal}{\emph{arXiv preprint arXiv:1811.00656}}
  (\bibinfo{year}{2018}).
\newblock


\bibitem[\protect\citeauthoryear{Li and Lyu}{Li and Lyu}{2019}]%
        {li2019exposing}
\bibfield{author}{\bibinfo{person}{Yuezun Li} {and} \bibinfo{person}{Siwei
  Lyu}.} \bibinfo{year}{2019}\natexlab{}.
\newblock \showarticletitle{Exposing DeepFake Videos By Detecting Face Warping
  Artifacts}. In \bibinfo{booktitle}{\emph{CVPRW}}.
\newblock


\bibitem[\protect\citeauthoryear{Li, Yang, Sun, Qi, and Lyu}{Li
  et~al\mbox{.}}{2020b}]%
        {li2020celeb}
\bibfield{author}{\bibinfo{person}{Yuezun Li}, \bibinfo{person}{Xin Yang},
  \bibinfo{person}{Pu Sun}, \bibinfo{person}{Honggang Qi}, {and}
  \bibinfo{person}{Siwei Lyu}.} \bibinfo{year}{2020}\natexlab{b}.
\newblock \showarticletitle{Celeb-df: A large-scale challenging dataset for
  deepfake forensics}. In \bibinfo{booktitle}{\emph{CVPR}}.
\newblock


\bibitem[\protect\citeauthoryear{Liu, Qi, and Torr}{Liu et~al\mbox{.}}{2020}]%
        {liu2020global}
\bibfield{author}{\bibinfo{person}{Zhengzhe Liu}, \bibinfo{person}{Xiaojuan
  Qi}, {and} \bibinfo{person}{Philip~HS Torr}.}
  \bibinfo{year}{2020}\natexlab{}.
\newblock \showarticletitle{Global texture enhancement for fake face detection
  in the wild}. In \bibinfo{booktitle}{\emph{CVPR}}.
\newblock


\bibitem[\protect\citeauthoryear{Masi, Killekar, Mascarenhas, Gurudatt, and
  AbdAlmageed}{Masi et~al\mbox{.}}{2020}]%
        {masi2020two}
\bibfield{author}{\bibinfo{person}{Iacopo Masi}, \bibinfo{person}{Aditya
  Killekar}, \bibinfo{person}{Royston~Marian Mascarenhas},
  \bibinfo{person}{Shenoy~Pratik Gurudatt}, {and} \bibinfo{person}{Wael
  AbdAlmageed}.} \bibinfo{year}{2020}\natexlab{}.
\newblock \showarticletitle{Two-branch recurrent network for isolating
  deepfakes in videos}. In \bibinfo{booktitle}{\emph{ECCV}}.
\newblock


\bibitem[\protect\citeauthoryear{Nagrani, Chung, and Zisserman}{Nagrani
  et~al\mbox{.}}{2017}]%
        {nagrani2017voxceleb}
\bibfield{author}{\bibinfo{person}{Arsha Nagrani}, \bibinfo{person}{Joon~Son
  Chung}, {and} \bibinfo{person}{Andrew Zisserman}.}
  \bibinfo{year}{2017}\natexlab{}.
\newblock \showarticletitle{Voxceleb: a large-scale speaker identification
  dataset}.
\newblock \bibinfo{journal}{\emph{arXiv preprint arXiv:1706.08612}}
  (\bibinfo{year}{2017}).
\newblock


\bibitem[\protect\citeauthoryear{Nazeri, Ng, Joseph, Qureshi, and
  Ebrahimi}{Nazeri et~al\mbox{.}}{2019}]%
        {nazeri2019edgeconnect}
\bibfield{author}{\bibinfo{person}{Kamyar Nazeri}, \bibinfo{person}{Eric Ng},
  \bibinfo{person}{Tony Joseph}, \bibinfo{person}{Faisal Qureshi}, {and}
  \bibinfo{person}{Mehran Ebrahimi}.} \bibinfo{year}{2019}\natexlab{}.
\newblock \showarticletitle{EdgeConnect: Structure Guided Image Inpainting
  using Edge Prediction}. In \bibinfo{booktitle}{\emph{ICCVW}}.
\newblock


\bibitem[\protect\citeauthoryear{Nguyen, Fang, Yamagishi, and Echizen}{Nguyen
  et~al\mbox{.}}{2019a}]%
        {nguyen2019multi}
\bibfield{author}{\bibinfo{person}{Huy~H Nguyen}, \bibinfo{person}{Fuming
  Fang}, \bibinfo{person}{Junichi Yamagishi}, {and} \bibinfo{person}{Isao
  Echizen}.} \bibinfo{year}{2019}\natexlab{a}.
\newblock \showarticletitle{Multi-task learning for detecting and segmenting
  manipulated facial images and videos}. In \bibinfo{booktitle}{\emph{BTAS}}.
\newblock


\bibitem[\protect\citeauthoryear{Nguyen, Yamagishi, and Echizen}{Nguyen
  et~al\mbox{.}}{2019b}]%
        {nguyen2019use}
\bibfield{author}{\bibinfo{person}{Huy~H Nguyen}, \bibinfo{person}{Junichi
  Yamagishi}, {and} \bibinfo{person}{Isao Echizen}.}
  \bibinfo{year}{2019}\natexlab{b}.
\newblock \showarticletitle{Use of a capsule network to detect fake images and
  videos}.
\newblock \bibinfo{journal}{\emph{arXiv preprint arXiv:1910.12467}}
  (\bibinfo{year}{2019}).
\newblock


\bibitem[\protect\citeauthoryear{Nirkin, Keller, and Hassner}{Nirkin
  et~al\mbox{.}}{2019}]%
        {nirkin2019fsgan}
\bibfield{author}{\bibinfo{person}{Yuval Nirkin}, \bibinfo{person}{Yosi
  Keller}, {and} \bibinfo{person}{Tal Hassner}.}
  \bibinfo{year}{2019}\natexlab{}.
\newblock \showarticletitle{Fsgan: Subject agnostic face swapping and
  reenactment}. In \bibinfo{booktitle}{\emph{ICCV}}.
\newblock


\bibitem[\protect\citeauthoryear{Pumarola, Agudo, Martinez, Sanfeliu, and
  Moreno-Noguer}{Pumarola et~al\mbox{.}}{2018}]%
        {pumarola2018ganimation}
\bibfield{author}{\bibinfo{person}{Albert Pumarola}, \bibinfo{person}{Antonio
  Agudo}, \bibinfo{person}{Aleix~M Martinez}, \bibinfo{person}{Alberto
  Sanfeliu}, {and} \bibinfo{person}{Francesc Moreno-Noguer}.}
  \bibinfo{year}{2018}\natexlab{}.
\newblock \showarticletitle{Ganimation: Anatomically-aware facial animation
  from a single image}. In \bibinfo{booktitle}{\emph{ECCV}}.
\newblock


\bibitem[\protect\citeauthoryear{Qian, Yin, Sheng, Chen, and Shao}{Qian
  et~al\mbox{.}}{2020}]%
        {qian2020thinking}
\bibfield{author}{\bibinfo{person}{Yuyang Qian}, \bibinfo{person}{Guojun Yin},
  \bibinfo{person}{Lu Sheng}, \bibinfo{person}{Zixuan Chen}, {and}
  \bibinfo{person}{Jing Shao}.} \bibinfo{year}{2020}\natexlab{}.
\newblock \showarticletitle{Thinking in frequency: Face forgery detection by
  mining frequency-aware clues}. In \bibinfo{booktitle}{\emph{ECCV}}.
\newblock


\bibitem[\protect\citeauthoryear{Qiu, Yao, and Mei}{Qiu et~al\mbox{.}}{2017}]%
        {qiu2017learning}
\bibfield{author}{\bibinfo{person}{Zhaofan Qiu}, \bibinfo{person}{Ting Yao},
  {and} \bibinfo{person}{Tao Mei}.} \bibinfo{year}{2017}\natexlab{}.
\newblock \showarticletitle{Learning spatio-temporal representation with
  pseudo-3d residual networks}. In \bibinfo{booktitle}{\emph{ICCV}}.
\newblock


\bibitem[\protect\citeauthoryear{Raffel, Shazeer, Roberts, Lee, Narang, Matena,
  Zhou, Li, and Liu}{Raffel et~al\mbox{.}}{2020}]%
        {raffel2019exploring}
\bibfield{author}{\bibinfo{person}{Colin Raffel}, \bibinfo{person}{Noam
  Shazeer}, \bibinfo{person}{Adam Roberts}, \bibinfo{person}{Katherine Lee},
  \bibinfo{person}{Sharan Narang}, \bibinfo{person}{Michael Matena},
  \bibinfo{person}{Yanqi Zhou}, \bibinfo{person}{Wei Li}, {and}
  \bibinfo{person}{Peter~J. Liu}.} \bibinfo{year}{2020}\natexlab{}.
\newblock \showarticletitle{Exploring the Limits of Transfer Learning with a
  Unified Text-to-Text Transformer}.
\newblock \bibinfo{journal}{\emph{JMLR}} (\bibinfo{year}{2020}).
\newblock


\bibitem[\protect\citeauthoryear{Rossler, Cozzolino, Verdoliva, Riess, Thies,
  and Nie{\ss}ner}{Rossler et~al\mbox{.}}{2019}]%
        {rossler2019faceforensics++}
\bibfield{author}{\bibinfo{person}{Andreas Rossler}, \bibinfo{person}{Davide
  Cozzolino}, \bibinfo{person}{Luisa Verdoliva}, \bibinfo{person}{Christian
  Riess}, \bibinfo{person}{Justus Thies}, {and} \bibinfo{person}{Matthias
  Nie{\ss}ner}.} \bibinfo{year}{2019}\natexlab{}.
\newblock \showarticletitle{Faceforensics++: Learning to detect manipulated
  facial images}. In \bibinfo{booktitle}{\emph{ICCV}}.
\newblock


\bibitem[\protect\citeauthoryear{Ruder, Dosovitskiy, and Brox}{Ruder
  et~al\mbox{.}}{2016}]%
        {ruder2016artistic}
\bibfield{author}{\bibinfo{person}{Manuel Ruder}, \bibinfo{person}{Alexey
  Dosovitskiy}, {and} \bibinfo{person}{Thomas Brox}.}
  \bibinfo{year}{2016}\natexlab{}.
\newblock \showarticletitle{Artistic style transfer for videos}. In
  \bibinfo{booktitle}{\emph{GCPR}}.
\newblock


\bibitem[\protect\citeauthoryear{Siarohin, Lathuili{\`e}re, Tulyakov, Ricci,
  and Sebe}{Siarohin et~al\mbox{.}}{2020}]%
        {siarohin2020first}
\bibfield{author}{\bibinfo{person}{Aliaksandr Siarohin},
  \bibinfo{person}{St{\'e}phane Lathuili{\`e}re}, \bibinfo{person}{Sergey
  Tulyakov}, \bibinfo{person}{Elisa Ricci}, {and} \bibinfo{person}{Nicu Sebe}.}
  \bibinfo{year}{2020}\natexlab{}.
\newblock \showarticletitle{First order motion model for image animation}.
\newblock \bibinfo{journal}{\emph{arXiv preprint arXiv:2003.00196}}
  (\bibinfo{year}{2020}).
\newblock


\bibitem[\protect\citeauthoryear{Simonyan and Zisserman}{Simonyan and
  Zisserman}{[n.d.]}]%
        {simonyan2014very}
\bibfield{author}{\bibinfo{person}{Karen Simonyan} {and}
  \bibinfo{person}{Andrew Zisserman}.} \bibinfo{year}{[n.d.]}\natexlab{}.
\newblock \showarticletitle{Very deep convolutional networks for large-scale
  image recognition}. In \bibinfo{booktitle}{\emph{ICLR}}.
\newblock


\bibitem[\protect\citeauthoryear{Sun, Yang, Liu, and Kautz}{Sun
  et~al\mbox{.}}{2018}]%
        {sun2018pwc}
\bibfield{author}{\bibinfo{person}{Deqing Sun}, \bibinfo{person}{Xiaodong
  Yang}, \bibinfo{person}{Ming-Yu Liu}, {and} \bibinfo{person}{Jan Kautz}.}
  \bibinfo{year}{2018}\natexlab{}.
\newblock \showarticletitle{Pwc-net: Cnns for optical flow using pyramid,
  warping, and cost volume}. In \bibinfo{booktitle}{\emph{CVPR}}.
\newblock


\bibitem[\protect\citeauthoryear{Suwajanakorn, Seitz, and
  Kemelmacher-Shlizerman}{Suwajanakorn et~al\mbox{.}}{2017}]%
        {suwajanakorn2017synthesizing}
\bibfield{author}{\bibinfo{person}{Supasorn Suwajanakorn},
  \bibinfo{person}{Steven~M Seitz}, {and} \bibinfo{person}{Ira
  Kemelmacher-Shlizerman}.} \bibinfo{year}{2017}\natexlab{}.
\newblock \showarticletitle{Synthesizing obama: learning lip sync from audio}.
\newblock \bibinfo{journal}{\emph{ACM TOG}} (\bibinfo{year}{2017}).
\newblock


\bibitem[\protect\citeauthoryear{Tan and Le}{Tan and Le}{2019}]%
        {tan2019efficientnet}
\bibfield{author}{\bibinfo{person}{Mingxing Tan} {and} \bibinfo{person}{Quoc
  Le}.} \bibinfo{year}{2019}\natexlab{}.
\newblock \showarticletitle{Efficientnet: Rethinking model scaling for
  convolutional neural networks}. In \bibinfo{booktitle}{\emph{ICML}}.
\newblock


\bibitem[\protect\citeauthoryear{Thies, Zollh{\"o}fer, and Nie{\ss}ner}{Thies
  et~al\mbox{.}}{2019}]%
        {thies2019deferred}
\bibfield{author}{\bibinfo{person}{Justus Thies}, \bibinfo{person}{Michael
  Zollh{\"o}fer}, {and} \bibinfo{person}{Matthias Nie{\ss}ner}.}
  \bibinfo{year}{2019}\natexlab{}.
\newblock \showarticletitle{Deferred neural rendering: Image synthesis using
  neural textures}.
\newblock \bibinfo{journal}{\emph{ACM TOG}} (\bibinfo{year}{2019}).
\newblock


\bibitem[\protect\citeauthoryear{Thies, Zollhofer, Stamminger, Theobalt, and
  Niessner}{Thies et~al\mbox{.}}{2016}]%
        {Thies_2016_CVPR}
\bibfield{author}{\bibinfo{person}{Justus Thies}, \bibinfo{person}{Michael
  Zollhofer}, \bibinfo{person}{Marc Stamminger}, \bibinfo{person}{Christian
  Theobalt}, {and} \bibinfo{person}{Matthias Niessner}.}
  \bibinfo{year}{2016}\natexlab{}.
\newblock \showarticletitle{Face2Face: Real-Time Face Capture and Reenactment
  of RGB Videos}. In \bibinfo{booktitle}{\emph{CVPR}}.
\newblock


\bibitem[\protect\citeauthoryear{Touvron, Cord, Douze, Massa, Sablayrolles, and
  J{\'e}gou}{Touvron et~al\mbox{.}}{2021}]%
        {touvron2020training}
\bibfield{author}{\bibinfo{person}{Hugo Touvron}, \bibinfo{person}{Matthieu
  Cord}, \bibinfo{person}{Matthijs Douze}, \bibinfo{person}{Francisco Massa},
  \bibinfo{person}{Alexandre Sablayrolles}, {and} \bibinfo{person}{Herv{\'e}
  J{\'e}gou}.} \bibinfo{year}{2021}\natexlab{}.
\newblock \showarticletitle{Training data-efficient image transformers \&
  distillation through attention}. In \bibinfo{booktitle}{\emph{ICML}}.
\newblock


\bibitem[\protect\citeauthoryear{Tripathy, Kannala, and Rahtu}{Tripathy
  et~al\mbox{.}}{2020}]%
        {tripathy2020icface}
\bibfield{author}{\bibinfo{person}{Soumya Tripathy}, \bibinfo{person}{Juho
  Kannala}, {and} \bibinfo{person}{Esa Rahtu}.}
  \bibinfo{year}{2020}\natexlab{}.
\newblock \showarticletitle{Icface: Interpretable and controllable face
  reenactment using gans}. In \bibinfo{booktitle}{\emph{WACV}}.
\newblock


\bibitem[\protect\citeauthoryear{Van~der Maaten and Hinton}{Van~der Maaten and
  Hinton}{2008}]%
        {van2008visualizing}
\bibfield{author}{\bibinfo{person}{Laurens Van~der Maaten} {and}
  \bibinfo{person}{Geoffrey Hinton}.} \bibinfo{year}{2008}\natexlab{}.
\newblock \showarticletitle{Visualizing data using t-SNE.}
\newblock \bibinfo{journal}{\emph{JMLR}} (\bibinfo{year}{2008}).
\newblock


\bibitem[\protect\citeauthoryear{Vaswani, Shazeer, Parmar, Uszkoreit, Jones,
  Gomez, Kaiser, and Polosukhin}{Vaswani et~al\mbox{.}}{2017}]%
        {vaswani2017attention}
\bibfield{author}{\bibinfo{person}{Ashish Vaswani}, \bibinfo{person}{Noam
  Shazeer}, \bibinfo{person}{Niki Parmar}, \bibinfo{person}{Jakob Uszkoreit},
  \bibinfo{person}{Llion Jones}, \bibinfo{person}{Aidan~N. Gomez},
  \bibinfo{person}{Lukasz Kaiser}, {and} \bibinfo{person}{Illia Polosukhin}.}
  \bibinfo{year}{2017}\natexlab{}.
\newblock \showarticletitle{Attention is All you Need}. In
  \bibinfo{booktitle}{\emph{NIPS}}.
\newblock


\bibitem[\protect\citeauthoryear{Wang, Wu, Chen, Han, Shrivastava, Lim, and
  Jiang}{Wang et~al\mbox{.}}{2022}]%
        {wang2022objectformer}
\bibfield{author}{\bibinfo{person}{Junke Wang}, \bibinfo{person}{Zuxuan Wu},
  \bibinfo{person}{Jingjing Chen}, \bibinfo{person}{Xintong Han},
  \bibinfo{person}{Abhinav Shrivastava}, \bibinfo{person}{Ser-Nam Lim}, {and}
  \bibinfo{person}{Yu-Gang Jiang}.} \bibinfo{year}{2022}\natexlab{}.
\newblock \showarticletitle{ObjectFormer for Image Manipulation Detection and
  Localization}. In \bibinfo{booktitle}{\emph{CVPR}}.
\newblock


\bibitem[\protect\citeauthoryear{Wang, Yang, Li, Wu, and Jiang}{Wang
  et~al\mbox{.}}{2021}]%
        {wang2021efficient}
\bibfield{author}{\bibinfo{person}{Junke Wang}, \bibinfo{person}{Xitong Yang},
  \bibinfo{person}{Hengduo Li}, \bibinfo{person}{Zuxuan Wu}, {and}
  \bibinfo{person}{Yu-Gang Jiang}.} \bibinfo{year}{2021}\natexlab{}.
\newblock \showarticletitle{Efficient Video Transformers with Spatial-Temporal
  Token Selection}.
\newblock \bibinfo{journal}{\emph{arXiv preprint arXiv:2111.11591}}
  (\bibinfo{year}{2021}).
\newblock


\bibitem[\protect\citeauthoryear{Wang, Wang, Zhang, Owens, and Efros}{Wang
  et~al\mbox{.}}{2020}]%
        {wang2020cnn}
\bibfield{author}{\bibinfo{person}{Sheng-Yu Wang}, \bibinfo{person}{Oliver
  Wang}, \bibinfo{person}{Richard Zhang}, \bibinfo{person}{Andrew Owens}, {and}
  \bibinfo{person}{Alexei~A Efros}.} \bibinfo{year}{2020}\natexlab{}.
\newblock \showarticletitle{CNN-generated images are surprisingly easy to
  spot... for now}. In \bibinfo{booktitle}{\emph{CVPR}}.
\newblock


\bibitem[\protect\citeauthoryear{Wodajo and Atnafu}{Wodajo and Atnafu}{2021}]%
        {wodajo2021deepfake}
\bibfield{author}{\bibinfo{person}{Deressa Wodajo} {and}
  \bibinfo{person}{Solomon Atnafu}.} \bibinfo{year}{2021}\natexlab{}.
\newblock \showarticletitle{Deepfake Video Detection Using Convolutional Vision
  Transformer}.
\newblock \bibinfo{journal}{\emph{arXiv preprint arXiv:2102.11126}}
  (\bibinfo{year}{2021}).
\newblock


\bibitem[\protect\citeauthoryear{Xiao, Dollar, Singh, Mintun, Darrell, and
  Girshick}{Xiao et~al\mbox{.}}{2021}]%
        {xiao2021early}
\bibfield{author}{\bibinfo{person}{Tete Xiao}, \bibinfo{person}{Piotr Dollar},
  \bibinfo{person}{Mannat Singh}, \bibinfo{person}{Eric Mintun},
  \bibinfo{person}{Trevor Darrell}, {and} \bibinfo{person}{Ross Girshick}.}
  \bibinfo{year}{2021}\natexlab{}.
\newblock \showarticletitle{Early Convolutions Help Transformers See Better}.
  In \bibinfo{booktitle}{\emph{NIPS}}.
\newblock


\bibitem[\protect\citeauthoryear{Xu, Das, and Saenko}{Xu et~al\mbox{.}}{2017}]%
        {xu2017r}
\bibfield{author}{\bibinfo{person}{Huijuan Xu}, \bibinfo{person}{Abir Das},
  {and} \bibinfo{person}{Kate Saenko}.} \bibinfo{year}{2017}\natexlab{}.
\newblock \showarticletitle{R-c3d: Region convolutional 3d network for temporal
  activity detection}. In \bibinfo{booktitle}{\emph{ICCV}}.
\newblock


\bibitem[\protect\citeauthoryear{Yang, Li, and Lyu}{Yang et~al\mbox{.}}{2019}]%
        {yang2019exposing}
\bibfield{author}{\bibinfo{person}{Xin Yang}, \bibinfo{person}{Yuezun Li},
  {and} \bibinfo{person}{Siwei Lyu}.} \bibinfo{year}{2019}\natexlab{}.
\newblock \showarticletitle{Exposing deep fakes using inconsistent head poses}.
  In \bibinfo{booktitle}{\emph{ICASSP}}.
\newblock


\bibitem[\protect\citeauthoryear{Yang and Guo}{Yang and Guo}{2020}]%
        {yang2019lafin}
\bibfield{author}{\bibinfo{person}{Yang Yang} {and} \bibinfo{person}{Xiaojie
  Guo}.} \bibinfo{year}{2020}\natexlab{}.
\newblock \showarticletitle{Generative Landmark Guided Face Inpainting}. In
  \bibinfo{booktitle}{\emph{PRCV}}.
\newblock


\bibitem[\protect\citeauthoryear{Zhang, Wu, Zhang, Zhu, Lin, Zhang, Sun, He,
  Mueller, Manmatha, et~al\mbox{.}}{Zhang et~al\mbox{.}}{2020}]%
        {zhang2020resnest}
\bibfield{author}{\bibinfo{person}{Hang Zhang}, \bibinfo{person}{Chongruo Wu},
  \bibinfo{person}{Zhongyue Zhang}, \bibinfo{person}{Yi Zhu},
  \bibinfo{person}{Haibin Lin}, \bibinfo{person}{Zhi Zhang},
  \bibinfo{person}{Yue Sun}, \bibinfo{person}{Tong He}, \bibinfo{person}{Jonas
  Mueller}, \bibinfo{person}{R Manmatha}, {et~al\mbox{.}}}
  \bibinfo{year}{2020}\natexlab{}.
\newblock \showarticletitle{Resnest: Split-attention networks}.
\newblock \bibinfo{journal}{\emph{arXiv preprint arXiv:2004.08955}}
  (\bibinfo{year}{2020}).
\newblock


\bibitem[\protect\citeauthoryear{Zhao, Zhou, Chen, Wei, Zhang, and Yu}{Zhao
  et~al\mbox{.}}{2021}]%
        {zhao2021multi}
\bibfield{author}{\bibinfo{person}{Hanqing Zhao}, \bibinfo{person}{Wenbo Zhou},
  \bibinfo{person}{Dongdong Chen}, \bibinfo{person}{Tianyi Wei},
  \bibinfo{person}{Weiming Zhang}, {and} \bibinfo{person}{Nenghai Yu}.}
  \bibinfo{year}{2021}\natexlab{}.
\newblock \showarticletitle{Multi-attentional Deepfake Detection}. In
  \bibinfo{booktitle}{\emph{CVPR}}.
\newblock


\bibitem[\protect\citeauthoryear{Zhou, Han, Morariu, and Davis}{Zhou
  et~al\mbox{.}}{2017}]%
        {zhou2017two}
\bibfield{author}{\bibinfo{person}{Peng Zhou}, \bibinfo{person}{Xintong Han},
  \bibinfo{person}{Vlad~I Morariu}, {and} \bibinfo{person}{Larry~S Davis}.}
  \bibinfo{year}{2017}\natexlab{}.
\newblock \showarticletitle{Two-stream neural networks for tampered face
  detection}. In \bibinfo{booktitle}{\emph{CVPRW}}.
\newblock


\bibitem[\protect\citeauthoryear{Zhu, Su, Lu, Li, Wang, and Dai}{Zhu
  et~al\mbox{.}}{2021}]%
        {zhu2020deformable}
\bibfield{author}{\bibinfo{person}{Xizhou Zhu}, \bibinfo{person}{Weijie Su},
  \bibinfo{person}{Lewei Lu}, \bibinfo{person}{Bin Li},
  \bibinfo{person}{Xiaogang Wang}, {and} \bibinfo{person}{Jifeng Dai}.}
  \bibinfo{year}{2021}\natexlab{}.
\newblock \showarticletitle{Deformable {\{}DETR{\}}: Deformable Transformers
  for End-to-End Object Detection}. In \bibinfo{booktitle}{\emph{ICLR}}.
\newblock


\bibitem[\protect\citeauthoryear{Zi, Chang, Chen, Ma, and Jiang}{Zi
  et~al\mbox{.}}{2020}]%
        {zi2020wilddeepfake}
\bibfield{author}{\bibinfo{person}{Bojia Zi}, \bibinfo{person}{Minghao Chang},
  \bibinfo{person}{Jingjing Chen}, \bibinfo{person}{Xingjun Ma}, {and}
  \bibinfo{person}{Yu-Gang Jiang}.} \bibinfo{year}{2020}\natexlab{}.
\newblock \showarticletitle{WildDeepfake: A Challenging Real-World Dataset for
  Deepfake Detection}. In \bibinfo{booktitle}{\emph{ACM MM}}.
\newblock


\end{thebibliography}

\end{document}